\documentclass[conference,9pt]{IEEEtran}
\IEEEoverridecommandlockouts
\usepackage{cite}
\usepackage{amsmath,amssymb,amsfonts}
\usepackage{caption}
\usepackage{subcaption}
\usepackage{enumerate}

\usepackage{algpseudocode}
\usepackage[linesnumbered,ruled,slide,boxed,vlined]{algorithm2e}
\SetEndCharOfAlgoLine{}
\algnewcommand\algorithmicforeach{\textbf{for each}}

\usepackage{graphicx}
\usepackage{textcomp}
\usepackage{xcolor}
\usepackage{multirow,multicol}
\graphicspath{{img/}}
\def\BibTeX{{\rm B\kern-.05em{\sc i\kern-.025em b}\kern-.08em
    T\kern-.1667em\lower.7ex\hbox{E}\kern-.125emX}}

\newcommand{\varSetX}{\mathbf{X}}
\newcommand{\varSetY}{\mathbf{Y}}

\newcommand{\subsetX}{\widetilde{\mathbf{X}}}
\newcommand{\combSet}{\mathbf{Q}}
\newcommand{\combSetSub}[1]{\mathbf{S}_{#1}}
\newcommand{\combSetPrime}[1]{\mathbf{P}_{#1}}
\newcommand{\true}{\mathbf{1}}
\newcommand{\false}{\mathbf{0}}
\newcommand{\var}[1]{v(#1)}

\newcommand{\vtree}{\mathbf{T}}
\newcommand{\univSet}{\mathbf{U}}
\newcommand{\negation}[1]{\bar{#1}}
\newcommand{\comp}[1]{\tilde{#1}}
\newcommand{\orthJoin}{\sqcup}
\newcommand{\emp}{\varepsilon}
\newcommand{\negempty}{\negation{\emp}}
\newcommand{\subtree}{\preccurlyeq}
\newcommand{\propSubtree}{\prec}
\newcommand{\set}[1]{\{ #1 \}}
\newcommand{\size}[1]{|#1|}
\newcommand{\bool}{\set{0, 1}}
\newcommand{\priVtree}[1]{pv(#1)}
\newcommand{\sndVtree}[1]{sv(#1)}

\newcommand{\trueAss}{\Sigma}
\newcommand{\assign}{\leftarrow}
\newcommand{\change}{{\tt Change}}

\newcommand{\stdSem}[1]{\langle #1 \rangle_s}
\newcommand{\zeroSem}[1]{\langle #1 \rangle_z}
\newcommand{\stdExtSem}[1]{\lVert #1 \rVert_s}
\newcommand{\zeroExtSem}[1]{\lVert #1 \rVert_z}

\newcommand{\eg}{{\it e.g.}}
\newcommand{\qed}{\hfill \square}

\newtheorem{theorem}{Theorem}
\newtheorem{definition}{Definition}

\newtheorem{example}{Example}

\begin{document}

\title{Variants of Tagged Sentential Decision Diagrams}

\author{\IEEEauthorblockN{Deyuan Zhong}
\IEEEauthorblockA{\textit{Department of Computer Science} \\
\textit{Jinan University}\\
Guangzhou, China \\
zhongdeyuan@stu2021.jnu.edu.cn}
\and
\IEEEauthorblockN{Mingwei Zhang}
\IEEEauthorblockA{\textit{Department of Computer Science} \\
\textit{Jinan University}\\
Guangzhou, China \\
mingweizhang@stu2022.jnu.edu.cn}
\and
\IEEEauthorblockN{Quanlong Guan}
\IEEEauthorblockA{\textit{Department of Computer Science} \\
\textit{Jinan University}\\
Guangzhou, China \\
guanql@jnu.edu.cn}
\and
\IEEEauthorblockN{Liangda Fang}
\IEEEauthorblockA{\textit{Department of Computer Science} \\
\textit{Jinan University}\\
Guangzhou, China \\
fangld@jnu.edu.cn}
\and
\IEEEauthorblockN{Zhaorong Lai}
\IEEEauthorblockA{\textit{Department of Computer Science} \\
\textit{Jinan University}\\
Guangzhou, China \\
laizhr@jnu.edu.cn}
\and
\IEEEauthorblockN{Yong Lai}
\IEEEauthorblockA{\textit{College of Computer Science and Technology} \\
\textit{JiLin University}\\
Changchun, China \\
laiy@jlu.edu.cn}
}

\maketitle

\begin{abstract}
A recently proposed canonical form of Boolean functions, namely tagged sentential decision diagrams (TSDDs), exploits both the standard and zero-suppressed trimming rules.
The standard ones minimize the size of sentential decision diagrams (SDDs) while the zero-suppressed trimming rules have the same objective as the standard ones but for zero-suppressed sentential decision diagrams (ZSDDs).
The original TSDDs, which we call zero-suppressed TSDDs (ZTSDDs), firstly fully utilize the zero-suppressed trimming rules, and then the standard ones.
In this paper, we present a variant of TSDDs which we call standard TSDDs (STSDDs) by reversing the order of trimming rules.
We then prove the canonicity of STSDDs and present the algorithms for binary operations on TSDDs.
In addition, we offer two kinds of implementations of STSDDs and ZTSDDs and acquire three variations of the original TSDDs.
Experimental evaluations demonstrate that the four versions of TSDDs have the size advantage over SDDs and ZSDDs.
\end{abstract}


\section{Introduction}
\looseness=-1
Knowledge compilation aims to transform a Boolean function into a tractable representation.
%Binary decision diagrams (BDDs) \cite{Bry1986} is one of the most notable representations that is widely employed for numerous fields of computer science including artificial intelligence \cite{ShiCD2019,KyrVZ2021}, computer-aided design \cite{ThiJE2022,MatM2022}, cryptography \cite{ZhangQTW2015,KniSM2020}, formal method \cite{MahD2021,WeiTJJ2022}.
Binary decision diagrams (BDDs) \cite{Bry1986} is one of the most notable representations that is widely employed for numerous fields of computer science including computer-aided design \cite{ThiJE2022,MatM2022}, cryptography \cite{ZhangQTW2015,KniSM2020}, formal method \cite{MahD2021,WeiTJJ2022}.
Interestingly, BDDs are a canonical form under the two restrictions: ordering and reduction, that means, any Boolean function has a unique BDD representation.
This property reduces the storage space of BDDs and enables an $O(1)$ time equality-test on BDDs.

\looseness=-1
Following the success of BDDs, a variant zero-suppressed BDDs (ZDDs) was proposed in \cite{Min1993}.
ZDDs enjoy the same properties: canonicity and supporting polytime Boolean operations as BDDs.
The main difference between BDDs and ZBDDs lies in their different reduction rules.
%As they use different reduction rules, their size significantly depends on the category of Boolean functions they represent. 
%BDDs are particularly suited for functions where adjacent input assignments have the same outcome (which we call \textit{homogeneous} Boolean functions), whereas ZBDDs are more compact for functions that often evaluate to $1$ when many variables are set to $0$ (which we call \textit{spare} Boolean functions).
%Some applications involve both these two classes of Boolean functions.
%\textcolor[rgb]{1,0,0}{
Some applications inspire several extensions of BDD that combine the reduction rules of BDDs and ZBDDs, including tagged BDDs (TBDDs) \cite{DijWM2017}, chain-reduced BDDs (CBDDs) \cite{Bry2020a}, chain-reduced ZDDs (CZDDs) \cite{Bry2020a} and edge-specified-reduction BDDs (ESRBDDs) \cite{BabJCM2022}.
%}
Thanks to the integration of two reduction rules, the above extensions are more compact representations than BDDs and ZDDs.

\looseness=-1
The theoretical foundation of BDDs is the Shannon decomposition \cite{Sha1938}, which splits a Boolean function into two subfunctions based on a single variable. 
Structured decomposition \cite{PipD2008}, an extension to the Shannon decomposition, splits a Boolean function according to a set of mutually exclusive subfunctions.
By using structured decomposition instead of the Shannon decomposition, a novel decision diagram, namely sentential decision diagram (SDD), was developed in \cite{Dar2011}.
Just as BDDs are characterized by a total order of variables, SDDs are characterized by a variable tree (vtree), that is, a full and binary tree whose leaves are variables.
%SDDs have the same interesting properties as BDDs: (1) they are a canoncial form w.r.t. a vtree under the trimming restriction, and (2) they support polytime Boolean operations.
%SDDs are a generalization of BDDs, since structured decomposition and vtree are extensions to the Shannon decomposition and variable order, respectively.
The advantage of SDDs over BDDs is providing a more succinct representation in theory and practice \cite{Bov2016,ChoD2013}.
In addition, \cite{NisYMN2016} proposed the zero-suppressed variant of SDDs (called ZSDDs), which is also based on structured decomposition, and applies the zero-suppressed trimming rules instead of the standard rules used in SDDs.
ZSDDs offer a more compact form for spare Boolean functions compared to SDDs.
In contrast, SDDs are more suitable for homogeneous Boolean functions.
In order to harness the relative strengths of SDDs and ZSDDs, \cite{FanFWZCY2019} designed a novel decision diagram, namely \textit{tagged SDDs (TSDDs)}, which combines the standard and zero-suppressed trimming rules.
%A SDD node consists of a vtree and a decomposition node which represents the structured decomposition.
%Each decomposition node is a list of pairs of two TSDD subnodes that represent the split subfunctions of the original Boolean functions.
%By adding an extra vtree to each SDD node, we acquire a TSDD node.
%To distinguish the two vtrees in a TSDD node, the extra vtree is referred to as the primary vtree and the original one as the secondary vtree.
%The TSDD nodes removed by the standard trimming rules is hidden in the primary and secondary vtrees while those nodes removed by the zero-suppressed trimming rules are implicitly represented by the decomposition node together with the secondary vtree.

\looseness=-1
In this paper, we investigate the variants of TSDDs.
To distinguish it from its variants, we call the original TSDD zero-suppressed TSDD (ZTSDD).
ZTSDD firstly fully utilizes the zero-suppressed trimming rules before adopting the standard ones.
By reversing the order of the trimming rules, we propose the first variant, namely standard TSDD (STSDD).
The syntactical definition of STSDD is the same as ZTSDD that is made up of two vtrees and a decomposition node.
However, STSDD uses the standard trimming rules as the first rule and the zero-suppressed ones as the second rule.
We also propose the semantics for STSDDs and design the trimming rules for STSDDs, obtaining the canonicity property of STSDDs.
In addition, we implement these two types of TSDDs in two ways: \textit{node-based} and \textit{edge-based}.
Basically, the node-based implementation specifies two vtrees and the decomposition node in a TSDD node.
In contrast, the edge-based implementation only keeps the secondary vtree in a TSDD node and associate the edge pointing to each STSDD subnode of the decomposition node with its primary vtree.
When a large number of nodes share the same secondary vtree and decomposable node, edge-based implementation utilizes less memory than node-based one.
Node-based implementation, on the other hand, uses less memory space to save TSDDs.
\cite{FanFWZCY2019} developed only edge-based implementation of ZTSDDs using C++ language.
Some critical data structures, such as unique table and cache table, were built directly on standard template library, making them less efficient.
We provide more efficient implementations of four TSDD variations by rewriting such data structures in C language.
We also compare SDDs and ZSDDs with the four TSDD variations in terms of size and compilation time of decision diagrams on an extensive set of benchmarks.
The experimental results support the effectiveness of our implementation and the relative compactness of TSDDs over SDDs and ZSDDs on the majority of test-cases.

\looseness=-1
The rest of this paper is organized as follows. 
Section 2 provides the preliminaries of Boolean function, combination set, the standard and zero-suppressed trimming rules. 
In Section 3, we give the syntax of TSDDs and two semantics for TSDDs, obtaining two versions of TSDDs: STSDDs and ZTSDDs.
We also design the compressness and trimming rules for STSDDs, gaining the canonicity property of STSDDs and offer two implementations for TSDDs.
In Section 4, we develop the algorithm for binary operations of combination sets on STSDDs.
Experimental evaluation for comparison among four variations of TSDDs with SDDs and ZSDDs appears in Section 5. 
Finally, Section 6 concludes this paper.

\section{Preliminaries}

\begin{figure*}
	%\vspace*{-1mm}
	\centering
	\includegraphics[width=\textwidth]{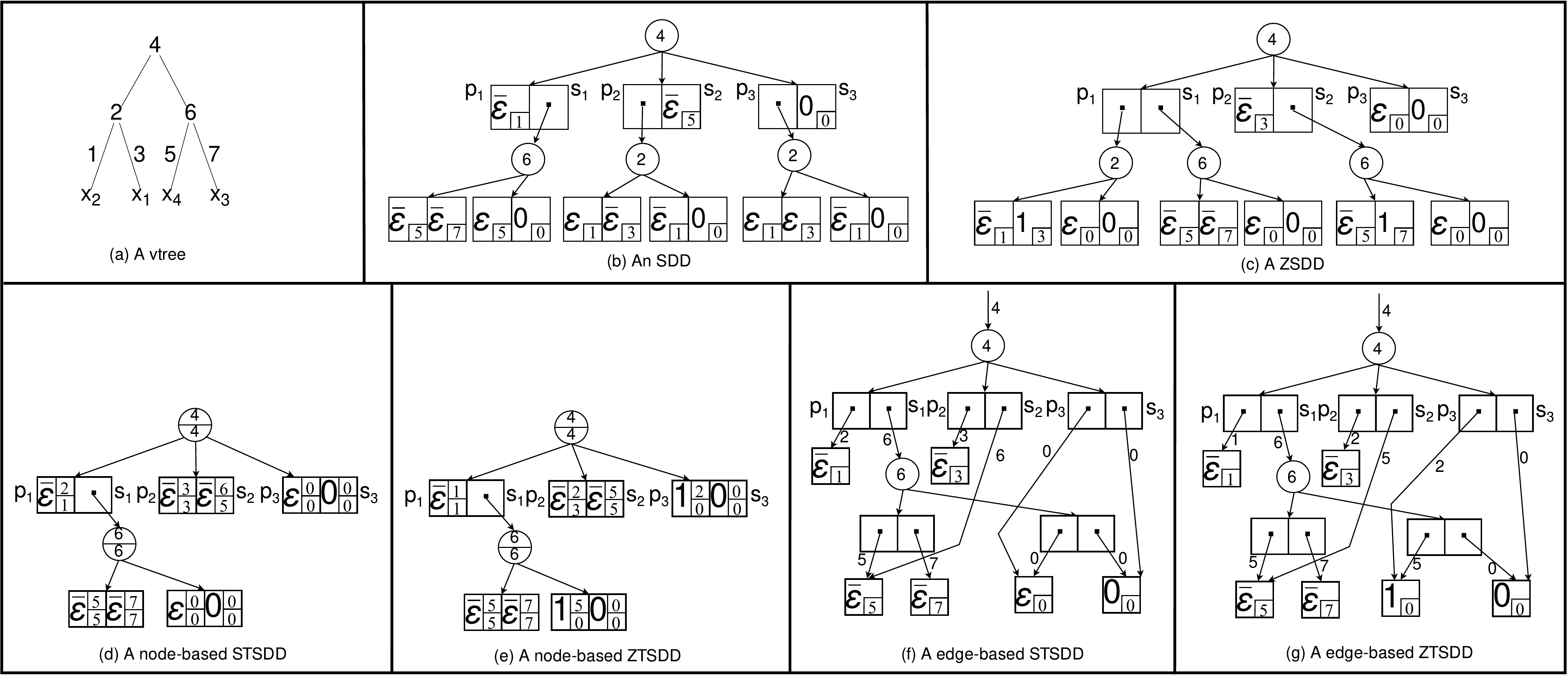}
		%\vspace*{-1mm}
	\caption{The vtree and the SDD, ZSDD and TSDD representations of the combination set $\set{\!\set{x_1,\! x_2,\! x_3,\! x_4},\! \set{x_2,\! x_3,\! x_4},\! \set{x_1,\! x_3,\! x_4},\! \set{x_1,\! x_4}\!}$.}
	%	\vspace*{-1mm}
	\label{fig:TSDD}
\end{figure*}

%\looseness=-1
Throughout this paper, we use lower case letters (\eg, $x_1, x_2$) for variables, and bold upper case letters (\eg, $\varSetX, \varSetY$) for sets of variables.
For a variable $x$, we use $\overline{x}$ to denote the negation of $x$. 
A \textit{literal} is a variable or a negated one. 
A \textit{truth assignment} over $\varSetX$ is a mapping $\sigma : \varSetX \mapsto \bool$. 
We let $\trueAss_{\varSetX}$ be the set of truth assignments over $\varSetX$. 
We say $f$ is a \textit{Boolean function} over $\varSetX$, which is a mapping: $\trueAss_{\varSetX} \mapsto \bool$. 
We use $\true$ (resp. $\false$) for the Boolean function that maps all assignments to $1$ (resp. $0$).
A \textit{combination} $\subsetX$ on $\varSetX$ is a subset of $\varSetX$.
Every combination $\subsetX$ corresponds to exactly one truth assignment $\sigma$, that is, $x \in \subsetX$ iff $\sigma(x) = 1$.
A \textit{combination set} $\combSet$ over $\varSetX$ is a collection of combinations on $\varSetX$.
It was shown that every combination set can be transformed into a Boolean function, and vice versa \cite{Min1993,NisYMN2016}.
The operations on combination sets include: union $\cup$, intersection $\cap$, difference $\setminus$, orthogonal join $\sqcup$ and change \cite{NisYMN2016}.
The definitions of the above operations are illustrated in Table \ref{tab:operation}.
We use $\univSet_{\varSetX}$ for the universe set of combinations on $\varSetX$.
For example, $\univSet_{\set{x_1, x_2}} = \set{\set{x_1, x_2}, \set{x_1}, \set{x_2}, \emptyset}$.
We remark that $\univSet_{\emptyset} = \set{\emptyset}$.

\begin{table}[t]
	\renewcommand{\arraystretch}{1.4}
	\smaller
	\centering
	\caption{Operations on combination sets.}
	\scalebox{1}{
		\begin{tabular}{|c|c|c|}  
			\hline
			Operation & Description & Definition \\
			\hline
			\hline
			$\combSet \cap \combSet'$ & intersection & $\set{\subsetX \mid \subsetX \in \combSet \text{ and } \subsetX \in \combSet'}$ \\
			\hline
			$\combSet \cup \combSet'$ & union & $\set{\subsetX \mid \subsetX \in \combSet \text{ or } \subsetX \in \combSet'}$ \\
			\hline
			$\combSet \setminus \combSet'$ & difference & $\set{\subsetX \mid \subsetX \in \combSet \text{ and } \subsetX \notin \combSet'}$ \\
			\hline
			$\combSet \orthJoin \combSet'$ & orthogonal join & $\set{\subsetX \cup \subsetX' \mid \subsetX \in \combSet \text{ and } \subsetX' \in \combSet'}$ \\
			\hline
			\multirow{2}{*}{$\change(\combSet, x)$} & \multirow{2}{*}{change} & $\set{\subsetX \cup \set{x} \mid \subsetX \in \combSet \text{ and } x \notin \subsetX} \cup$ \\
			& & \hspace*{-2mm}$\set{\subsetX \setminus \set{x} \mid \subsetX \in \combSet \text{ and } x \in \subsetX}$ \\
			\hline
		\end{tabular}
	}
	\label{tab:operation}
\end{table}

%\begin{table}[t]
%	\renewcommand{\arraystretch}{1.4}
%	\smaller
%	\centering
%	\caption{Operations on combination sets.}
%	\scalebox{1}{
%		\begin{tabular}{|c|c|c|}  
%			\hline
%			符号表示 & 操作名字 & 定义 \\
%			\hline
%			\hline
%			$\combSet \cap \combSet'$ & 交集 & $\set{\subsetX \mid \subsetX \in \combSet \text{ 和 } \subsetX \in \combSet'}$ \\
%			\hline
%			$\combSet \cup \combSet'$ & 并集 & $\set{\subsetX \mid \subsetX \in \combSet \text{ 或 } \subsetX \in \combSet'}$ \\
%			\hline
%			$\combSet \setminus \combSet'$ & 差集 & $\set{\subsetX \mid \subsetX \in \combSet \text{ 和 } \subsetX \notin \combSet'}$ \\
%			\hline
%			$\combSet \orthJoin \combSet'$ & 正交 & $\set{\subsetX \cup \subsetX' \mid \subsetX \in \combSet \text{ 和 } \subsetX' \in \combSet'}$ \\
%			\hline
%			\multirow{2}{*}{$\change(\combSet, x)$} & \multirow{2}{*}{变换} & $\set{\subsetX \cup \set{x} \mid \subsetX \in \combSet \text{ 和 } x \notin \subsetX} \cup$ \\
%			& & \hspace*{-2mm}$\set{\subsetX \setminus \set{x} \mid \subsetX \in \combSet \text{ 和 } x \in \subsetX}$ \\
%			\hline
%		\end{tabular}
%	}
%	\label{tab:operation1}
%\end{table}

Let $\varSetX$ and $\varSetY$ be two disjoint and non-empty sets of variables.
We say the set $\set{(\combSetPrime{1}, \combSetSub{1}), \cdots, (\combSetPrime{n}, \combSetSub{n})}$ is an $(\varSetX, \varSetY)$-\textit{decomposition} of a combination set $\combSet$, iff $\combSet = [\combSetPrime{1} \orthJoin \combSetSub{1}] \cup \cdots \cup[\combSetPrime{n} \orthJoin \combSetSub{n}]$ where every $\combSetPrime{i}$ (resp. $\combSetSub{i}$) is a combination set over $\varSetX$ (resp. $\varSetY$).
%, and both $\varSetX$ and $\varSetY$ contain at least one variable.
%Every pair $(\semantic{p_i}, \semantic{s_i})$ is called an \textit{element} of this decomposition. 
%For each element, $p_i$ is called a \textit{prime} and $s_i$ is called a \textit{sub}. % \cite{NisYMN2016}. 
%For a prime $p_i$, we say $s_i$ is the sub w.r.t. $p_i$. 
A decomposition is \textit{compressed} iff $\combSetSub{i} \neq \combSetSub{j}$ for $i \neq j$.
An $(\varSetX, \varSetY)$-decomposition is called an $(\varSetX, \varSetY)$-\textit{partition}, iff (1) every $\combSetPrime{i}$ is non-empty, (2) $\combSetPrime{i} \cap \combSetPrime{j} = \emptyset$ for $i \neq j$, and (3) $\combSetPrime{i} \cup \cdots \cup \combSetPrime{n} = \univSet_{\varSetX}$.
%}

%A combination set can be represented as a diagram by recursively applying $(\varSetX, \varSetY)$-partitions (shown in Fig.\ref{fig:DD}) until we can't apply $(\varSetX, \varSetY)$-partitions on it(both $\varSetX$ and $\varSetY$ contain at least one variable). 
%We use $\combSet,p_i,s_i$ to denote decision diagrams, and use $\combSet(\varSetX, \varSetY),p_i(\varSetX),s_i(\varSetY)$ to denote corresponding combination sets.
%Besides, $p_i$ is called a \textit{prime} and $s_i$ is called a \textit{sub}. % \cite{NisYMN2016}.

%\looseness=-1
A \textit{vtree} is a full binary tree whose leaves are labeled by variables, which generalizes variable orders.
%A decision diagram is based on the notion of vtrees, which generalize variable orders, and . 
For a vtree $\vtree$, we use $\var{\vtree}$ for the set of variables appearing in leaves of $\vtree$, and $\vtree_l$ and $\vtree_r$ for the left and right subtrees of $\vtree$ respectively.
There is a special leaf node labeled by $0$ that can be considered as a child of any vtree node and $\var{0} = \emptyset$.
The notation $\vtree^1 \subtree \vtree^2$ denotes that $\vtree^1$ is a subtree of $\vtree^2$ and $\vtree^1 \propSubtree \vtree^2$ means that $\vtree^1$ is a proper subtree.

Based the notion of vtrees, a combination set can be graphically represented by a \textit{structured decomposable diagram}\cite{FanFWZCY2019}.
\begin{definition} %[\cite{FanFWZCY2019}]
	%Let $\vtree$ be a vtree.
	A structured decomposable diagram is a pair $(\vtree, \alpha)$ where $\vtree$ is a vtree and $\alpha$ is recursively defined as follows
	\begin{itemize}
		\item $\alpha$ is a terminal node labeled by one of the four symbols: $\true$, $\false$, $\emp$ and $\negempty$, and $\vtree$ is any vtree.
		\item $\alpha$ is a decomposition node $\set{(p_1, s_1), \cdots, (p_n, s_n)}$
		satisfying the following conditions:
		\begin{enumerate}
			\item each $p_i$ is a structured decomposable diagram $(\vtree_i^1, \beta)$ where $\vtree_i^1 \subtree \vtree_l$;
			\item each $s_i$ is a structured decomposable diagram $(\vtree_i^2, \gamma)$ where $\vtree_i^2 \subtree \vtree_r$.
		\end{enumerate}
	\end{itemize}
	%The size of $\alpha$, denoted $|\alpha|$, is obtained by summing the sizes of all its decompositions. 
	%The notation $\emp$ denotes the conjunction of negative literals, and $\negempty$ denotes the negation of $\emp$.
\end{definition}

\looseness=-1
Every pair $(p_i, s_i)$ of a decomposition node is called an \textit{element} where $p_i$ is called a \textit{prime} and $s_i$ is called a \textit{sub}. % \cite{NisYMN2016}. 
%For a prime $p_i$, we say $s_i$ is the sub w.r.t. $p_i$.
%Every pair (pi, si) is called an element of this decomposition.

We hereafter provide two ways to interpret a structured decomposable diagram $(\vtree^2, \alpha)$ as a combination set, which we call \textit{standard} and \textit{zero-suppressed semantics}.
%Under the standard semantic
%\textcolor[rgb]{1,0,0}{
Since the standard semantics depends on an extra vtree $\vtree^1$, it is a mapping from structured decomposable diagrams and vtrees into combination sets.
%}

\begin{definition} \label{def:stdSem}
	Let $\vtree^1$ be a vtree and $(\vtree^2, \alpha)$ be a structured decomposable diagram where $\vtree^2 \subtree \vtree^1$.
	The \textit{standard semantics} $\stdSem{\vtree^1, (\vtree^2, \alpha)}$ is recursively defined as follows:
	\begin{itemize}
		\item $\stdSem{\vtree^1, (\vtree^2, \true)} = \univSet_{\vtree^1}$ and $\stdSem{\vtree^1, (\vtree^2, \false)} = \emptyset$;
		\item $\stdSem{\vtree^1, (\vtree^2, \emp)} = \univSet_{\var{\vtree^1} \setminus \var{\vtree^2}}$ and $\stdSem{\vtree^1, (\vtree^2, \negempty)} = \univSet_{\var{\vtree^1} \setminus \var{\vtree^2}} \orthJoin (\univSet_{\var{\vtree^2}} \setminus \set{\emptyset})$;
		\item $\stdSem{\vtree^1, (\vtree^2, \set{(p_1, s_1), \cdots, (p_n, s_n)})} = \univSet_{\var{\vtree^1} \setminus \var{\vtree^2}} \orthJoin \left[\bigcup\limits_{i = 1}\limits^{n} (\stdSem{\vtree_l^2,p_i} \orthJoin \stdSem{\vtree_r^2,s_i}) \right]$.
	\end{itemize}	
%	The \textit{zero-suppressed semantic} is recursively defined as:
%	\begin{itemize}
%		\item $\zeroSem{\vtree^1, (\vtree^2, \true)} = \univSet_{\var{\vtree^2}}$ and $\zeroSem{\vtree^1, (\vtree^2, \false)} = \emptyset$;
%		
%		
%		\item $\zeroSem{\vtree^1, (\vtree^2, \emp)} = \set{\emptyset}$ and $\zeroSem{\vtree^1, (\vtree^2, \negempty)} = \univSet_{\var{\vtree^2}} \setminus \set{\emptyset}$;
%		
%		\item $\zeroSem{(\vtree^{1}, \vtree^{2}, \set{(p_1, s_1), \cdots, (p_n, s_n)})} = \bigcup \limits_{i = 1} \limits^{n} (\zeroSem{p_i} \orthJoin \zeroSem{s_i})$.
%	\end{itemize}
\end{definition}

%\textcolor[rgb]{1,0,0}{
The standard semantics $\stdSem{\vtree^1, (\vtree^2, \alpha)}$ contains two combination sets.
%}
The \textit{main combination set} is based on $\vtree^2$ and $\alpha$.
The four terminal nodes $\true$, $\false$, $\emp$ and $\negempty$ represent $\univSet_{\var{\vtree^2}}$, $\emptyset$, $\set{\emptyset}$ and $\univSet_{\var{\vtree^2}} \setminus \set{\emptyset}$, respectively.
The decomposition node $\set{(p_1, s_1), \cdots, (p_n, s_n)}$ denotes the combination set that is the union of the orthogonal join of $\stdSem{p_i}$ and $\stdSem{s_i}$ for every pair $(p_i, s_i)$.
% $\bigcup\limits_{i = 1}\limits^{n} (\stdSem{p_i} \orthJoin \stdSem{s_i})$.
The \textit{auxiliary combination set} is the universe set over $\var{\vtree^1} \setminus \var{\vtree^2}$.
The standard semantics is the orthogonal join of main and auxiliary combination sets.
For example, the combination set of $(\vtree^{1}, \vtree^{2}, \true)$ is $\univSet_{\var{\vtree^2}} \orthJoin \univSet_{\var{\vtree^1} \setminus \var{\vtree^2}}$, and hence being $\univSet_{\var{\vtree^1}}$.

The zero-suppressed semantics $\zeroSem{\vtree^1, (\vtree^2, \alpha)}$ is only the main combination set, which can be easily defined.
For example, $\zeroSem{\vtree^1, (\vtree^2, \set{(p_1, s_1), \cdots, (p_n, s_n)})} = \bigcup\limits_{i = 1}\limits^{n} (\stdSem{p_i} \orthJoin \stdSem{s_i})$.
%The extra vtree $\vtree^1$ is not required for the zero-suppressed semantics.
%\textcolor[rgb]{1,0,0}{
We introduce the extra vtree $\vtree^1$ in the zero-suppressed semantics in accordance with the standard semantics though it is not required for the zero-suppressed semantics.
%}
%Due to space limitation, we do not provide 

Based on the standard semantics, we impose some restrictions on structured decomposable diagram and obtain the definition of sentenial decision diagram (SDD).
\begin{definition}\label{def:SDD}
	A structured decomposable diagram $(\vtree, \alpha)$ is a sentenial decision diagram, if one of the following holds:
	\begin{enumerate}
		\item $\alpha$ is a terminal node labeled by $\true$ or $\false$, and $\vtree = 0$.
		\item $\alpha$ is a terminal node labeled by $\emp$ or $\negempty$, and $\vtree$ is a leaf node.
		\item $\alpha$ is a decomposition node $\set{(p_1, s_1), \cdots, (p_n, s_n)}$, and all of the following hold:
		\begin{itemize}
			\item $\stdSem{\vtree_l, p_i} \neq \emptyset$ for $1 \leq i \leq n$;
			\item $\stdSem{\vtree_l, p_i} \cap \stdSem{\vtree_l, p_j} = \emptyset$ for $i \neq j$;
			\item $\bigcup \limits_{i = 1}^n \stdSem{\vtree_l, p_i} = \univSet_{\var{\vtree_l}}$.
		\end{itemize}
	\end{enumerate}
\end{definition}

The definition of zero-suppressed sentenial decision diagram (ZSDD) is the same as SDD, except that 
(1) we require $\vtree$ to be the special vtree $0$ when $\alpha$ is a terminal node labeled by $\emp$;
(2) the vtree $\vtree$ can be any leaf node when $\alpha$ is labeled by $\true$;
(3) we use the zero-suppressed semantics for the decomposition node. % $\{(p_1, s_1), \cdots, (p_n, s_n)\}$.

%\begin{definition}\label{def:ZSDD}
%	A structured decomposable diagram $(\vtree, \alpha)$ is a zero-suppressed sentenial decision diagram, if one of the following holds:
%	\begin{enumerate}
%		\item $\alpha$ is a terminal node labeled by $\false$ or $\emp$, and $\vtree = 0$.
%		\item $\alpha$ is a terminal node labeled by $\true$ or $\negempty$, and $\vtree$ is a leaf node.
%		\item $\alpha$ is a decomposition node $\{(p_1, s_1), \cdots, (p_n, s_n)\}$, and all of the following hold:
%		\begin{itemize}
%			\item $\zeroSem{\vtree_l, p_i} \neq \false$ for $1 \leq i \leq n$;
%			\item $\zeroSem{\vtree_l, p_i} \land \zeroSem{\vtree_l, p_j} = \false$ for $i \neq j$;
%			\item $\bigvee \limits_{i = 1}^n \zeroSem{\vtree_l, p_i} = \true$.
%		\end{itemize}
%	\end{enumerate}
%\end{definition}

An SDD can be transformed to an equivalent one with smaller size by the following the standard compressness and  trimming rules.
\begin{itemize}
	\looseness=-1
	\item Standard compression rule (S-compression rule): if $\stdSem{\vtree_r, s_i} = \stdSem{\vtree_r, s_j}$, then replace $(\vtree, \{(p_1, s_1), \cdots, (p_i, s_i), \\ \cdots, (p_j, s_j), \cdots, (p_n, s_n)\})$ with $(\vtree, \{(p_1, s_1), \cdots, (p', s_i), \\ \cdots, (p_n, s_n)\})$ where $\stdSem{\vtree_l, p'}  = \stdSem{\vtree_l, p_i} \cup \stdSem{\vtree_l, p_j}$.
	
	\item Standard trimming rule (S-trimming rule): 
	\begin{enumerate}[(a)]
		\item replace the diagram $(\vtree,\set{(p, s)})$ by the diagram $s$.
		\item if $\stdSem{\vtree_r,s_1} = \univSet_{\var{\vtree_r}}$ and $\stdSem{\vtree_r,s_2} = \emptyset$, then replace the diagram $(\vtree, \set{(p_1, s_1), (p_2, s_2)})$ by the diagram $p_1$.
	\end{enumerate}
\end{itemize}

\looseness=-1
The S-compression rule combines two elements $(p_i, s_i)$ and $(p_j, s_j)$ when $s_i$ and $s_j$ denotes the same combination set.
The two S-trimming rules aim to remove the universe set over a subset of variables in a decomposition node.
By repeatedly applying the S-compression and trimming rules, we can create the unique SDD.

Similarly, we can define the zero-suppressed compression and trimming rules for ZSDD. % by using the zero-suppressed semantics instead.
\begin{itemize}
	\looseness=-1
	\item Zero-suppressed compression rule (Z-compression rule): if $\zeroSem{\vtree_r, s_i} = \zeroSem{\vtree_r, s_j}$, then replace $(\vtree, \{(p_1, s_1), \cdots, (p_i, s_i), \\ \cdots, (p_j, s_j), \cdots, (p_n, s_n)\})$ with $(\vtree, \{(p_1, s_1), \cdots, (p', s_i), \\ \cdots, (p_n, s_n)\})$ where $\zeroSem{\vtree_l, p'}  = \zeroSem{\vtree_l, p_i} \cup \zeroSem{\vtree_l, p_j}$.
	
	\item Zero-suppressed trimming rule (Z-trimming rule):
	\begin{enumerate}[(a)]
		\item if $\zeroSem{\vtree_l,p_1} = \set{\emptyset}$ (resp. $\zeroSem{\vtree_r,s_1} = \set{\emptyset}$) and $\zeroSem{\vtree_r,s_2} = \emptyset$, then replace the diagram $(\vtree,\set{(p_1, s_1), (p_2, s_2)})$ by the diagram $s_1$ (resp. $p_1$);
		\item if $\zeroSem{\vtree_r,s} = \emptyset$, then replace the diagram $(\vtree,\set{(p, s)})$ by the diagram $s$;
	\end{enumerate}
\end{itemize}

The Z-compression rule is similar to the S-compression rule except that it uses the zero-suppressed semantics.
The two Z-trimming rules seek to eliminate $\set{\emptyset}$.
We acquire the canonical representation via utilizing the Z-trimming rule on ZSDDs.

\begin{example}
	\looseness=-1
	Fig. \ref{fig:TSDD}(a) shows the vtree $\vtree$ where its left subtree $\vtree_l$ involves $\varSetX: \set{x_1, x_2}$ while its right one $\vtree_r$ involves $\varSetY: \set{x_3, x_4}$.
	Fig. \ref{fig:TSDD}(b) depicts an SDD representing the combination set $\combSet = \set{\set{x_1, x_2, x_3, x_4}, \set{x_2, x_3, x_4}, \set{x_1, x_3, x_4}, \set{x_1,x_4}}$ based on $\vtree$.
	The $(\varSetX, \varSetY)$-partition of $\combSet$ contains three elements: $(\overbrace{\set{\set{x_1, x_2}, \set{x_2}}}^{\combSetPrime{1}}, \overbrace{\set{\set{x_3, x_4}}}^{\combSetSub{1}})$, $(\overbrace{\set{\set{x_1}}}^{\combSetPrime{2}}, \overbrace{\set{\set{x_3, x_4}, \set{x_4}}}^{\combSetSub{2}})$ and $(\overbrace{\set{\emptyset}}^{\combSetPrime{3}}, \overbrace{\emptyset}^{\combSetSub{3}})$.
	Each combination subset $\combSetPrime{i}$ (resp. $\combSetSub{i}$) corresponds to the node $p_i$ (resp. $s_i$) of the SDD.
	The ZSDD representation for $\combSet$ with smaller nodes than the SDD one is shown in Fig. \ref{fig:TSDD}(c). $\qed$
\end{example}

\section{Tagged Sentential Decision Diagrams}
In this section, we will first provide a general structure, namely extended structured decomposable diagram (ESDD), that is, the syntactic definition for TSDDs, and then with two different semantics, that are, a mapping from ESDDs to combination sets.
The ESDD with the standard semantics is called standard TSDD (STSDD) while it is called zero-suppressed TSDD (ZTSDD) under the zero-suppressed semantics.
We also present the trimming rules for STSDD so as to reduce the size of STSDD and obtain the canonicity theorem of STSDDs.
Finally, we provide two implementations for TSDDs: node-based and edge-based.
Hence, we obtain four versions of TSDDs.

\subsection{The Syntax and semantics}
In order to facilitate combining two types of trimming rules, we first provide a general structure, namely extended structured decomposable diagram (ESDD).

%\begin{figure}
%	\smaller
%	\centering
%	\includegraphics[width=0.4\textwidth]{DecisionDiagram.pdf}
%	\caption{Decision Diagram}
%	\vspace*{-3mm}
%	\label{fig:DD}
%\end{figure}

\begin{figure}[t]
	\begin{subfigure}[b]{0.24\textwidth}
		\centering
		\includegraphics[scale=0.3]{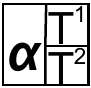}
		\caption{A terminal node}
%		\label{fig:Level_swap_single_variable}
	\end{subfigure}
	\begin{subfigure}[b]{0.24\textwidth}
		\centering
		\includegraphics[scale=0.3]{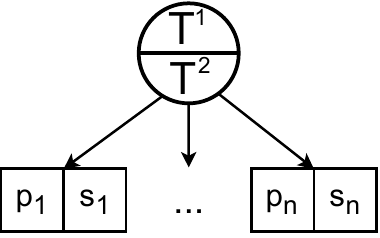}
		\caption{A decomposition node}
%		\label{fig:Level_swap_double_variable}
	\end{subfigure}
	\caption{Two types of nodes of ESDDs}
	\label{fig:ESDD}
\end{figure}

%\begin{figure}[t]
%	\begin{subfigure}[b]{0.24\textwidth}
%		\centering
%		\includegraphics[scale=0.3]{TerminalNode.pdf}
%		\caption{终端结点}
%		%		\label{fig:Level_swap_single_variable}
%	\end{subfigure}
%	\begin{subfigure}[b]{0.24\textwidth}
%		\centering
%		\includegraphics[scale=0.3]{DecomposedNode.pdf}
%		\caption{分解结点}
%		%		\label{fig:Level_swap_double_variable}
%	\end{subfigure}
%	\caption{Two types of nodes of ESDDs}
%	\label{fig:ESDD1}
%\end{figure}

\begin{definition}
	An ESDD is a tuple $(\vtree^{1}, \vtree^{2}, \alpha)$ s.t. $\vtree^{2} \subtree \vtree^{1}$, which is recursively defined as:
	\begin{itemize}
		\item $\alpha$ is a terminal node labeled by one of the four symbols: $\true$, $\false$, $\emp$ and $\negempty$; %\textcolor[rgb]{1,0,0}{ (shown in Fig. \ref{fig:DD}(b))}.
		
		\item $\alpha$ is a decomposition node $\set{(p_1, s_1), \cdots, (p_n,s_n)}$ satisfying the following:
		
		\begin{itemize}
			\item each $p_i$ is an ESDD $(\vtree^3, \vtree^4, \beta)$ where $\vtree^4 \subtree \vtree^3 \propSubtree \vtree^2$;
			\item each $s_i$ is an ESDD $(\vtree^5, \vtree^6, \gamma)$ where $\vtree^6 \subtree \vtree^5 \propSubtree \vtree^2$.
		\end{itemize}
	%	\textcolor[rgb]{1,0,0}{We use a box likes $\framebox[0.4cm]{}\framebox[0.4cm]{}$ to denote an element(shown in Fig. \ref{fig:DD}(a)).}
	\end{itemize}
\end{definition}

%Given an ESDD $(\vtree^{1}, \vtree^{2}, \alpha)$, the three components
%When $\alpha$ is a terminal node, 

An ESDD $F = (\vtree^{1}, \vtree^{2}, \alpha)$ consists of three components: the primary vtree $\vtree^{1}$, the secondary vtree $\vtree^{2}$ and the terminal (or decomposition) node $\alpha$.
As seen in Fig. \ref{fig:ESDD}(a), when $\alpha$ is a terminal node, the above three components are represented by a square where $\alpha$ is shown in the left side of the square, $\vtree^{1}$ in the upper-right corner and $\vtree^{2}$ in the lower-right corner.
%As shown in Fig. \ref{fig:ESDD}(b), 
When $\alpha$ is a decomposition node, the primary and secondary vtrees are displayed as a circle with outgoing edges pointing to the elements as shown in Fig. \ref{fig:ESDD}(b).
Each element $(p_i, s_i)$ is represented by a paired box where the left box represents the prime $p_i$ and the right box stands for the sub $s_i$.
We use $\priVtree{F}$ for the primary vtree of $F$ and $\sndVtree{F}$ for the secondary vtree. % and $\node{F}$ for the node.
The size of $\alpha$, denoted by $\size{\alpha}$, is the sizes of all of its decompositions.

To interpret ESDDs, we provide the semantics, that is, a mapping from ESDDs into combination sets.

\begin{definition} \label{def:stdExtSem}
	Let $(\vtree^{1}, \vtree^{2}, \alpha)$ be an ESDD.
	The \textit{standard semantics} $\stdExtSem{(\vtree^{1}, \vtree^{2}, \alpha)}$ is recursively defined as:
	\begin{itemize}
		\item $\stdExtSem{(\vtree^{1}, \vtree^{2}, \true)} = \univSet_{\var{\vtree^1}}$ and $\stdExtSem{(\vtree^{1}, \vtree^{2}, \false)} = \emptyset$;		
		
		\item $\stdExtSem{(\vtree^{1}, \vtree^{2}, \emp)} = \univSet_{\var{\vtree^1} \setminus \var{\vtree^2}}$ and $\stdExtSem{(\vtree^{1}, \vtree^{2}, \negempty)} = \univSet_{\var{\vtree^1} \setminus \var{\vtree^2}} \orthJoin (\univSet_{\var{\vtree^2}} \setminus \set{\emptyset})$;
		
		\item $\stdExtSem{(\vtree^{1}, \vtree^{2}, \set{(p_1, s_1), \cdots, (p_n, s_n)})} = \univSet_{\var{\vtree^1} \setminus \var{\vtree^2}} \orthJoin \left[\bigcup\limits_{i = 1}\limits^{n} (\stdExtSem{p_i} \orthJoin \stdExtSem{s_i}) \right]$.
	\end{itemize}
\end{definition}

Since every ESDD involves an extra vtree $\vtree^1$ compared to structured decision diagrams, the standard semantics for ESDDs is similar to structured decision diagrams (cf. Definition \ref{def:stdSem}).

%Let $F$ be an ESDD $(\vtree^{1}, \vtree^{2}, \alpha)$.
%The semantics $\stdExtSem{F}$ contains two combination sets. 
%The \textit{main combination set} of $F$ is based on $\vtree^2$ and $\alpha$.
%The four terminal nodes $\true$, $\false$, $\emp$ and $\negempty$ represents $\univSet_{\var{\vtree^2}}$, $\emptyset$, $\set{\emptyset}$ and $\univSet_{\var{\vtree^2}} \setminus \set{\emptyset}$, respectively.
%The decomposition node $\set{(p_1, s_1), \cdots, (p_n, s_n)}$ denotes the combination set $\bigcup\limits_{i = 1}\limits^{n} (\stdExtSem{p_i} \orthJoin \stdExtSem{s_i})$.
%The \textit{auxiliary combination set} of $F$ is the universe set over $\var{\vtree^1} \setminus \var{\vtree^2}$.
%The semantics $\stdExtSem{F}$ is the orthogonal join of main and auxiliary combination sets.
%For example, the combination set of $(\vtree^{1}, \vtree^{2}, \true)$ is $\univSet_{\var{\vtree^2}} \orthJoin \univSet_{\var{\vtree^1} \setminus \var{\vtree^2}}$, and hence being $\univSet_{\var{\vtree^1}}$.

%A decomposable node $\set{(p_1, s_1), \cdots, (p_n, s_n)}$ is a \textit{standard partition}, if (1) $\stdExtSem{p_i} \neq \emptyset$ for $1 \leq i \leq n$; (2) $\stdExtSem{p_i} \cap \stdExtSem{p_j} = \emptyset$ for $ i\neq j$; and (3) $\bigcup \limits_{i = 1} \limits^{n} \stdExtSem{p_i} = \univSet_{\var{\vtree^2_l}}$.

%Based on the standard semantic, we provide 
A standard tagged sentential decision diagram (STSDD) is an ESDD with the following constraints.
\begin{definition} \label{def:synSTSDD}
	An ESDD $(\vtree^1, \vtree^2, \alpha)$ is an STSDD, if one of the following holds:
	\begin{itemize}
		\item $\alpha$ is a terminal node labeled by $\false$ and $\vtree^1 = \vtree^2 = 0$.
		\item $\alpha$ is a terminal node labeled by $\emp$ and $\vtree^2 = 0$.
		\item $\alpha$ is a terminal node labeled by $\negempty$ and $\vtree^2$ is a leaf node.
		\item $\alpha$ is a decomposable node $\set{(p_1, s_1), \cdots, (p_n, s_n)}$ and $\set{(\stdExtSem{p_1}, \stdExtSem{s_1}), \cdots, (\stdExtSem{p_n}, \stdExtSem{s_n})}$ is an $(\varSetX, \varSetY)$-partition where $\varSetX = \var{\vtree^2_l}$ and $\varSetY = \var{\vtree^2_r}$.
	\end{itemize}
\end{definition}

We remark that we use the terminal node $\emp$ instead of $\true$ in STSDDs since $\stdExtSem{(\vtree^1, \vtree^2, \true)} = \stdExtSem{(\vtree^1, 0, \emp)}$ for any vtrees $\vtree^1$ and $\vtree^2$.

\subsection{Canonicity}
We hereafter design the standard tagged compression and trimming rules for reducing the size of STSDD and obtaining the canonicity property of STSDDs.

\begin{figure*}
	\centering
	\includegraphics[width=\textwidth]{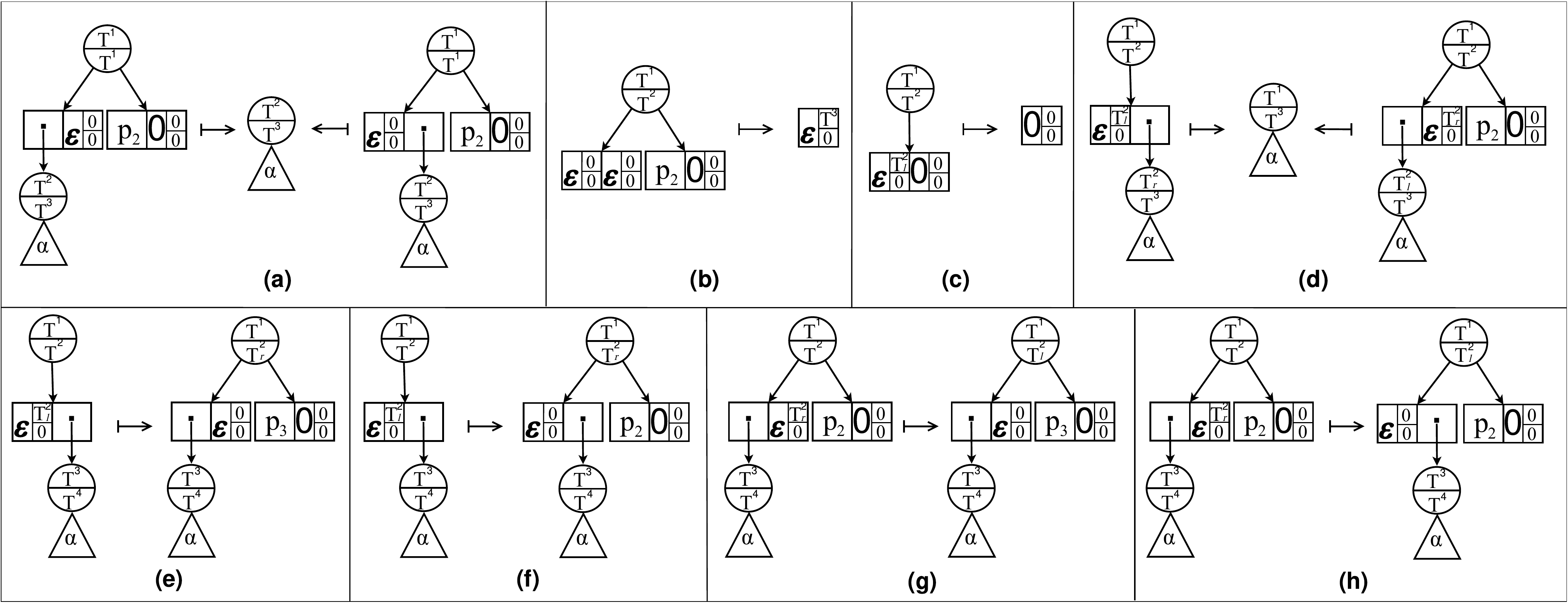}
	%	\vspace*{-3mm}
	\caption{Trimming rules for STSDD}
	\vspace*{-3mm}
	\label{fig:trimming}
\end{figure*}

\begin{itemize}
	\item Standard tagged compression rule (ST-compression rule): \\
	if $\stdExtSem{s_i} = \stdExtSem{s_j}$, then replace
	$(\vtree^1, \vtree^2, \set{(p_1, s_1), \cdots, (p_i, s_i), \cdots, (p_j, s_j), \cdots, (p_n, s_n)})$ with $(\vtree^1, \vtree^2, \set{ (p_1,s_1), \cdots, (p_i',s_i), \cdots, (p_n, s_n) } )$ where $\stdExtSem{p_i'} = \stdExtSem{p_i} \cup \stdExtSem{p_j}$.
	
	\item Standard tagged trimming rule (ST-trimming rule) (Fig \ref{fig:trimming}): %there are totally nine trimming rules, and they are shown in Fig. \ref{fig:trimming}.
%	We take (d) for example: if $p = (\vtree^2_l, 0, \emp)$ and $s = (\vtree^2_r, \vtree^3, \alpha)$, then replace $(\vtree^1, \vtree^2, \set{(p, s)})$ with $(\vtree^1, \vtree^3, \alpha)$;}
	
	\begin{enumerate}[(a)]
		\item if $p_1 = (\vtree^2, \vtree^3, \alpha)$, $\stdExtSem{s_1} = \set{\emptyset}$ and $\stdExtSem{s_2} = \emptyset$, or $\stdExtSem{p_1} = \set{\emptyset}$, $s_1 = (\vtree^2, \vtree^3, \alpha)$ and $\stdExtSem{s_2} = \emptyset$, then replace $(\vtree^1,\vtree^1, \set{(p_1,s_1),(p_2,s_2)})$ with $(\vtree^2, \vtree^3, \alpha)$;
		
		\item if $\stdExtSem{p_1} = \stdExtSem{s_1} = \set{\emptyset}$, $\stdExtSem{s_2} = \emptyset$ and $\vtree^2$ is $\vtree^1_l$ or $\vtree^1_r$, then replace $(\vtree^1, \vtree^2, \{(p_1,s_1),$ $(p_2,s_2)\})$ with $(\vtree^3, 0, \emp)$, where $\vtree^3 \!=\! \vtree^1_l$ when $\vtree^2 \!=\! \vtree^1_l$ and $\vtree^3 \!=\! \vtree^1_r$ when $\vtree^2 \!=\! \vtree^1_r$.
		
		 %(resp. $\vtree^1_r$). % if $\vtree^2 = \vtree^1_l$ (resp. $\vtree^1_r$);
		
		\item if $p = (\vtree^2_l, 0, \emp)$ and $\stdExtSem{s} = \emptyset$, then replace $(\vtree^1, \vtree^2, \set{(p, s)})$ with $(0, 0, \false)$.
		
		\item if $p_1 = (\vtree^2_l, \vtree^3, \alpha)$, $s_1 = (\vtree_r^2, 0, \emp)$ and $\stdExtSem{s_2} = \emptyset$ (resp. \\ $p = (\vtree^2_l, 0, \emp)$ and $s = (\vtree^2_r, \vtree^3, \alpha)$), then replace $(\vtree^1, \vtree^2_r, \set{((0, 0, \emp), s_1), (p_2, (0, 0, \false))})$  (resp. $(\vtree^1, \vtree^2, \set{(p, s)})$) with $(\vtree^1, \vtree^3, \alpha)$;
		
		\item if $p_1 = (\vtree^2_l, 0, \emp)$, $s_1 = (\vtree^3, \vtree^4, \alpha)$ and $\vtree^3 \subtree (\vtree^2_r)_l$, then \\ replace $(\vtree^1, \vtree^2, \set{(p_1, s_1)})$ with $(\vtree^1, \vtree^2_r, \set{(s_1, (0, 0, \emp)), \\ (p_2, (0,0,\false))})$ where $ \stdExtSem{p_2} = \univSet_{(\vtree^2_r)_l} \setminus \stdExtSem{s_1}$;
		
		\item if $p_1 = (\vtree^2_l, 0, \emp)$, $s_1 = (\vtree^3, \vtree^4, \alpha)$ and $\vtree^3 \subtree (\vtree^2_r)_r$, then \\ replace $(\vtree^1, \vtree^2, \set{(p_1,s_1)})$ with $(\vtree^1, \vtree^2_r, \set{((0, 0, \emp), s_1), \\ (p_2, (0,0,\false))})$ where $\stdExtSem{p_2} = \univSet_{(\vtree^2_r)_l} \setminus \set{\emptyset}$;
		
		%\item if $p_1 = (\vtree^2_l, \vtree^3, \alpha)$, $s_1 = (\vtree_r^2, 0, \emp)$ and $\stdExtSem{s_2} = \emptyset$, then replce $(\vtree^1, \vtree^2, $ $\set{(p_1, s_1), (p_2, s_2)})$ with $(\vtree^1, \vtree^3, \alpha)$;
		
		\item if $p_1 = (\vtree^3, \vtree^4, \alpha)$, $s_1 = (\vtree_r^2, 0, \emp)$, $\stdExtSem{s_2} = \emptyset$ and $\vtree^3 \subtree (\vtree^2_l)_l$,then replace $(\vtree^1, \vtree^2, \set{(p_1, s_1), (p_2, s_2)})$ with $(\vtree^1,\! \vtree_l^2,\! {(p_1, (0, 0, \emp)),\! (p_3,\! s_2)})$ where $\!\stdExtSem{p_3} \!=\! \univSet_{(\vtree^2_l)_l} \!\setminus \stdExtSem{p_1}$;
		
		\item if $p_1 = (\vtree^3, \vtree^4, \alpha)$, $s_1 = (\vtree_r^2, 0, \emp)$, $\stdExtSem{s_2} = \emptyset$ and $\vtree^3 \subtree (\vtree^2_l)_r$, then replace $(\vtree^1, \vtree^2, \set{(p_1, s_1), (p_2, s_2)})$ with $(\vtree^1,\! \vtree^2_l,\! \set{((0, 0, \emp),\! p_1), (p_3,\! s_2)})$ where $\!\stdExtSem{p_3} \!=\! \univSet_{(\vtree^2_l)_l} \!\setminus\! \set{\emptyset}$.
	\end{enumerate}
\end{itemize}

The goal of ST-compression rule is to combine elements with the same subs.
The ST-trimming rules are shown in Fig. \ref{fig:trimming}.
Rules (a) and (b) are used to eliminate the sub-diagram representing the set $\set{\emptyset}$ whereas rules (c) -- (h) aim to reduce the sub-diagram denoting the universe set over a subset of variables.
A STSDD is compressed (resp. trimmed), if no ST-compression (resp. trimming) rule can be applied in it. 
We hereafter state the important property of compressed and trimmed STSDDs.

\begin{theorem}	\label{thm:canonicity}
	Given a vtree $\vtree$ over $\varSetX$, for any combination set $\combSet$ over $\varSetX$, there is a unique compressed and trimmed STSDD $(\vtree^1, \vtree^2, \alpha)$ s.t. $\vtree^1 \subtree \vtree$ and  $\stdExtSem{(\vtree^1, \vtree^2, \alpha)} = \combSet$.
\end{theorem}

\looseness=-1
%(??)
%Thanks to the additional vtree in an ESDD node and the above trimming rules, STSDD has compactness advantages over both SDD and ZSDD.
Thanks to the additional vtree and the above trimming rules, STSDD has compactness advantages over both SDD and ZSDD.

\begin{example}
	We continue to Example 1.
	The combination set $\combSet$ in SDD, ZSDD and STSDD representations are shown in Fig. \ref{fig:TSDD}(b) -- (d), respectively. 
	The combination subsets $\combSetPrime{1}$, $\combSetSub{2}$ and $\combSetSub{3}$ can be represented as terminal nodes in SDD while $\combSetPrime{2}$, $\combSetPrime{3}$ and $\combSetSub{3}$ can be in ZSDD.
	Hence, the combination set has SDD and ZSDD representations of size $9$.
	All of the above $5$ combination subsets are represented by terminal nodes in STSDD.
	The STSDD representation, in comparison, is only $5$ in size smaller than SDD and ZSDD. $\qed$
\end{example}

\subsection{Zero-suppressed Variant}
%\textcolor[rgb]{1,0,0}{
In a STSDD $(\vtree^1, \vtree^2, \alpha)$, S-trimming rules are applied from the primary vtree $\vtree^1$ to the secondary one $\vtree^2$ and Z-trimming rules are applied from the secondary vtree $\vtree^2$ to the primary vtree of each of the terminal node $\alpha$, or the prime $p_i$ and the sub $s_i$ of the decomposition node $\alpha$.
%}
%洺玮觉得这句话表意不清
We hereafter define a variant of STSDD by reversing the order of trimming rules, that is, Z-trimming rules are implied first and S-trimming rules second.

\begin{definition} \label{def:zeroExtSem}
	Let $(\vtree^{1}, \vtree^{2}, \alpha)$ be an ESDD.
	The \textit{zero-suppressed semantics} $\zeroExtSem{(\vtree^{1}, \vtree^{2}, \alpha)}$ is recursively defined as:
	\begin{itemize}
		\item $\zeroExtSem{(\vtree^{1}, \vtree^{2}, \true)} = \univSet_{\var{\vtree^2}}$ and $\zeroExtSem{(\vtree^{1}, \vtree^{2}, \false)} = \emptyset$;

		\item $\zeroExtSem{(\vtree^{1}, \vtree^{2}, \emp)} = \set{\emptyset}$ and $\zeroExtSem{(\vtree^{1}, \vtree^{2}, \negempty)} = \univSet_{\var{\vtree^2}} \setminus \set{\emptyset}$;
		
		\item $\zeroExtSem{(\vtree^{1}, \vtree^{2}, \set{(p_1, s_1), \cdots, (p_n, s_n)})} = \\ \bigcup\limits_{i = 1}\limits^{n} \left[ \univSet_{\var{\vtree^2} \setminus (pv(p_i) \cup pv(s_i))} \orthJoin \zeroExtSem{p_i} \orthJoin \zeroExtSem{s_i} \right]$.
	\end{itemize}
\end{definition}

When $\alpha$ is the terminal node, the zero-suppressed semantics is only the main combination set of $\alpha$.
When $\alpha$ is the decomposition node, besides the main combination set, the zero-suppressed semantics contains an extra combination set, that is, the universal set of  $\var{\vtree^2} \setminus (pv(p_i) \cup pv(s_i))$ for each element $(p_i, s_i)$.

%A decomposable node $\set{(p_1, s_1), \cdots\!, (p_n, s_n)}$ is a \textit{zero-suppressed partition}, if (1) $\zeroExtSem{p_i} \!\neq\! \emptyset$ for $1 \leq i \leq n$; (2) $\zeroExtSem{p_i} \cap \zeroExtSem{p_j} \!=\! \emptyset$ for $i \neq j$; and (3) $\bigcup \limits_{i = 1} \limits^{n} \zeroExtSem{p_i} \!=\! \univSet_{\var{\vtree^2_l}}$.

Based on the zero-suppressed semantics, we provide the zero-suppressed variant of TSDD, namely zero-suppressed TSDD (ZTSDD).

\begin{definition}
	An ESDD $(\vtree^1, \vtree^2, \alpha)$ is an ZTSDD, if one of the following holds:
	%	A ZTSDD $(\vtree^1, \vtree^2, \alpha)$ is an ESDD that satisfies the following:
	\begin{itemize}
		\item $\alpha$ is a terminal node labeled by $\false$ and $\vtree^1 = \vtree^2 = 0$.
		\item $\alpha$ is a terminal node labeled by $\true$ and $\vtree^2=0$.
		\item $\alpha$ is a terminal node labeled by $\negempty$ and $\vtree^2$ is a leaf node.
		\item $\alpha$ is a decomposable node $\set{(p_1, s_1), \cdots, (p_n, s_n)}$ and $\set{(\zeroExtSem{p_1}, \zeroExtSem{s_1}), \cdots, (\zeroExtSem{p_n}, \zeroExtSem{s_n})}$ is an $(\varSetX, \varSetY)$-partition where $\varSetX = \var{\vtree^2_l}$ and $\varSetY = \var{\vtree^2_r}$.
	\end{itemize}
\end{definition}

We remark that the terminal node $\emp$ is omitted in ZTSDD due to the fact that $\zeroExtSem{(\vtree^1, \vtree^2, \emp)} = \zeroExtSem{(\vtree^1, 0, \true)}$ for any vtrees $\vtree^1$ and $\vtree^2$.
In addition, ZTSDD is a canonical form for combination set by applying zero-suppressed tagged compression and trimming rules.
%\textcolor[rgb]{1,0,0}{
%Due to space limit, we present the compression and trimming rules for ZTSDDs and the proof of canonicity theorem in Appendix.
%}

Fig. \ref{fig:TSDD}(e) shows the ZTSDD for representing the example.
Since ZTSDD both enjoy the advantages of SDD and ZSDD, it has smaller size $5$ than SDD and ZSDD, which is the same as STSDD.
We remark that ZTSDDs and STSDDs in general have different sizes for representing the same combination set given the same vtree.

\begin{figure}
	\centering
	\begin{subfigure}[b]{0.5\textwidth}
	\centering
	\includegraphics[scale=0.3]{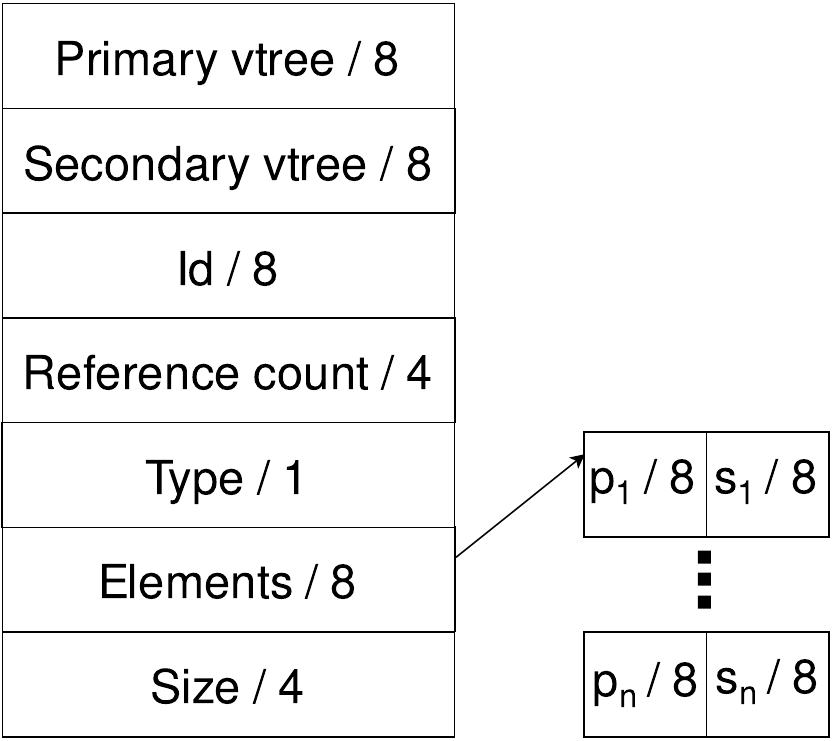}
	\caption{Node-based TSDD node}
	%		\label{fig:Level_swap_single_variable}
	\end{subfigure}
	\begin{subfigure}[b]{0.5\textwidth}
	\centering
	\includegraphics[scale=0.3]{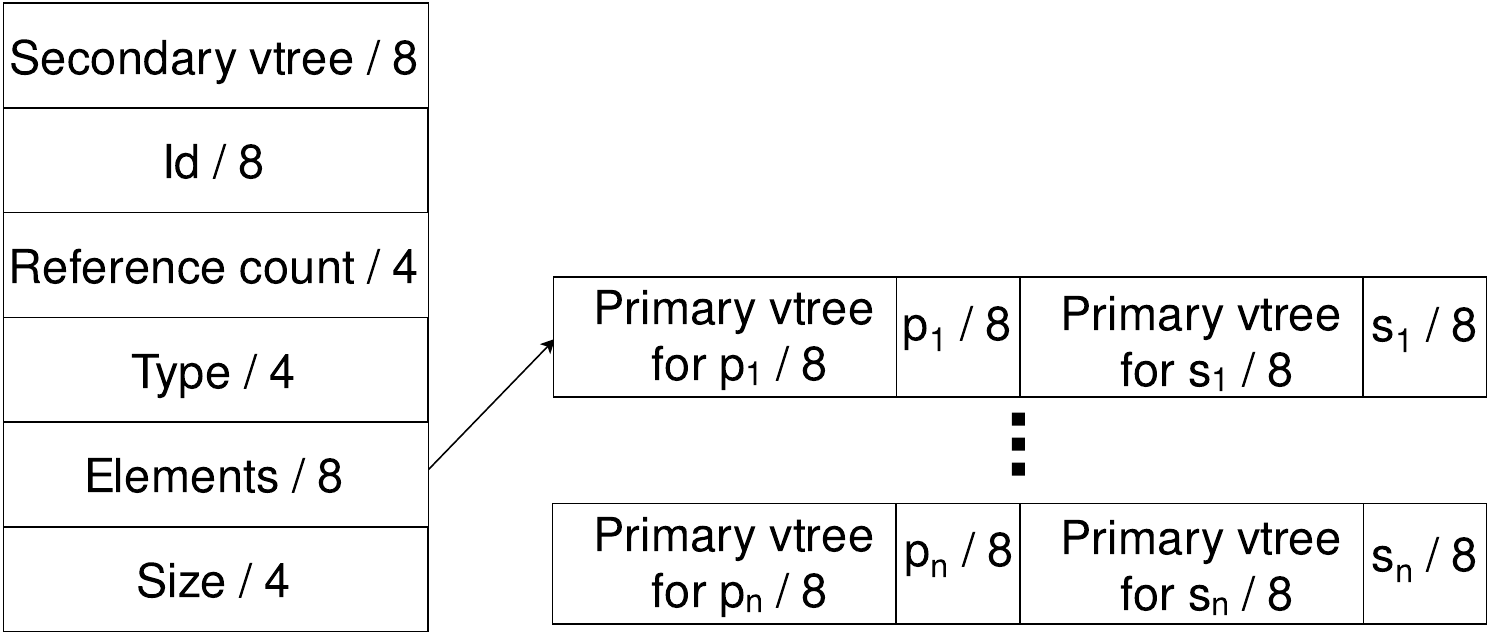}
	\caption{Edge-based TSDD node}
	%		\label{fig:Level_swap_double_variable}
	\end{subfigure}
	\caption{Two implementations for TSDDs}
	\label{fig:implementation}
\end{figure}

\subsection{Edge-based Variant}
\looseness=-1
In Definition \ref{def:synSTSDD}, both the primary vtree $\vtree^1$ and the secondary one $\vtree^2$ are kept in each TSDD node. 
Such TSDD node is called node-based TSDD node.
We now introduce the edge-based variant of TSDD.
The main distinctions between node-based and edge-based TSDD are:
(1) Each edge-based TSDD node only includes the secondary vtree $\vtree^2$ and the terminal/decomposition node $\alpha$;
(2) Each element of a decomposition node consists of not only the prime $p$ and the sub $s$ but also two vtrees $\vtree_p$ and $\vtree_s$ that are the primary vtrees of $p$ and $s$, respectively;
(3) There is an extra edge pointing to the root node denoting the primary vtree of the root node.

\looseness=-1
The main data structures of node-based TSDDs and edge-based TSDDs are shown in Fig. \ref{fig:implementation}, respectively.
%We explain each data item as follows.
Suppose that the TSDD node has $n$ pairs of primes and subs.
We remark that $n = 0$ when the node is a terminal node.
In the node-based TSDDs, each node requires at least $41 + 16n$ bytes: $8$ bytes for the pointer to the primary vtree, $8$ bytes for the pointer to the secondary vtree, $8$ bytes for the id of the node in the unique table that ensures no two equivalent TSDD nodes are stored, $4$ bytes for the reference count that is used to garbage collection, $1$ byte for the type of this node: terminal node or decomposition node, $8$ bytes for the pointer to a singly-linked list of pairs of primes nd subs, and $4$ bytes for the number of elements.
The node of edge-based TSDD has the similar data structure with node-based TSDD. 
However, each edge-based TSDD node do not have the primary vtree and each element has two additional primary vtrees for the prime and the sub.
The size of a edge-based node is $33 + 32n$.

%\textcolor[rgb]{1,0,0}{
%The main data structures of node-based ESDDs and edge-based ESDDs for $64$-bit computer are shown in Fig. \ref{fig_implement}, respectively, and the number after $/$ in each box indicates the number of bytes occupied by this field.
%%We discuss each data item as follows.
%%In the node-based TSDDs, the \textit{primary vtree} ($8$ bytes) and the \textit{secondary vtree} ($8$ bytes) are the pointers to vtree nodes.
%%\textit{Hash link} ($8$ bytes) is an index of the TSDD node, which ensures no two equivalent TSDD nodes are stored in the unique table.
%%\textit{Reference count} ($4$ bytes) records the number of reference of the TSDD node that is used to garbage collection.
%%\textit{Type} ($1$ byte) indicates the type of this node.
%%\textit{Size} ($4$ bytes) is the number $n$ of elements when this TSDD node is a decomposition node.
%%When it is a terminal node, the size $n$ is $0$.
%%\textit{Elements} ($8$ bytes) is a union data.
%%If the TSDD node is decomposition node, then it is a pointer to a list of pairs of primes $p_i$ and subs $s_i$.
%%Otherwise, it is one of the four cases of terminal nodes: $\true$, $\false$, $\emp$ and $\overline{\emp}$.
%We can see that the size of a node-based node is $41 + 16n$.
%The node of edge-based TSDD has the similar data structure with node-based TSDD. 
%However, each edge-based TSDD node do not have the primary vtree and each element has two additional primary vtrees for the prime and the sub.
%The size of a edge-based node is $33 + 32n$.
%}

%(??)
Fig. \ref{fig:TSDD}(f) and (g) show the edge-based STSDD and ZTSDD representations for the same example.
%We make a comparison between node-based STSDD representation (Fig. \ref{fig:TSDD}-(d)) and edge-based one on the prime $p_4$ and the sub $s_2$.
%The node-based STSDD represents $p_4$ and $s_2$ by a node with the following members: the primary vtree, the secondary vtree and the terminal node while edge-based STSDD by an edge denoting the primary vtree together with the node that consists of only the secondary vtree and the terminal node.
%Edge-based STSDD representation allow different edges to point to the same node.
%For example, $p_4$ and $s_2$ have different primary vtrees $5$ and $6$, but with the same secondary vtree $5$ and the terminal node $\negempty$.
%Hence, in edge-based STSDD, $p_4$ and $s_2$ are represented by two different edges pointing to the same node.
Node-based STSDD representation for $\combSet$ needs $449$ bytes whereas that of edge-based one requires $432$ bytes.
Due to this sharing mechanism, edge-based variant consumes less memory than node-based one when numerous nodes share the same secondary vtree and terminal (or decomposable) node.
Otherwise, node-based variant is a data structure occupying less memory space for storing TSDDs.

\section{Operations on STSDD}
\looseness=-1
In this section, we will design the algorithms of STSDDs for achieving the five operations on combination sets.
We first introduce a normalization algorithm that serves as the basis of the above algorithms, and then introduce a unified algorithm that accomplishes the three operations: intersection, union and difference, and followed by two algorithms for orthogonal join and change operations.

\subsection{Apply}
The normalization rules can be considered as a reverse of trimming rules.
There are two types of normalization rules.
Given a vtree $\vtree^3$ s.t. $\vtree^1 \subtree \vtree^3$, the first one is to transform a STSDD $(\vtree^1, \vtree^2, \alpha)$ into an equivalent one $(\vtree^3, \vtree^4, \beta)$, denoted by ${\tt Normalize1}((\vtree^1, \vtree^2, \alpha), \vtree^3)$.
\begin{enumerate}[(a)]
	\item if $\vtree^1 = \vtree^3$, then $\vtree^4 = \vtree^2$ and $\beta = \alpha$. 
	
	\item if $\vtree^1 \propSubtree \vtree^3_l$, then $\vtree^4 = \vtree^3$ and $\beta = \set{((\vtree^1, \vtree^2, \alpha), (0 ,0, \emp)),\\  (p, (0, 0, \false))}$ where $\stdExtSem{p} = \univSet_{\vtree^3_l} \setminus \stdExtSem{(\vtree^1, \vtree^2, \alpha)}$.
	
	\item if $\vtree^1 \propSubtree \vtree^3_r$, then $\vtree^4 = \vtree^3$ and $\beta = \set{((0, 0, \emp), (\vtree^1, \vtree^2, \alpha)),\\ (p, (0, 0, \false))}$ where $\stdExtSem{p} = \univSet_{\vtree^3_l} \setminus \set{\emptyset}$.
\end{enumerate}

\looseness=-1
The second type of normalization rules takes a STSDD $(\vtree^1, \vtree^2, \alpha)$ and a vtree $\vtree^4$ where $\vtree^2 \subtree \vtree^4 \subtree \vtree^1$ as input, and outputs the resulting STSDD, denoted by ${\tt Normalize2}((\vtree^1, \vtree^2, \alpha), \vtree^4)$, with the same combination set as $(\vtree^1, \vtree^2, \alpha)$.
\begin{enumerate}[(a)]
	\item if $\vtree^2 = \vtree^4$, then $\vtree^3 = \vtree^1$ and $\beta = \alpha$.
	
	\item if $\vtree^2 \propSubtree \vtree^4_l$, then $\vtree^3 = \vtree^1$ and $\beta = \set{((\vtree^4_l, \vtree^2, \alpha), (\vtree^4_r, 0, \emp)),\\ (p, (0, 0, \false))}$ where $\stdExtSem{p} = \univSet_{\vtree^4_l} \setminus \stdExtSem{(\vtree^1, \vtree^2, \alpha)}$.
	
	\item if $\vtree^2 \!\propSubtree\! \vtree^4_r$, then $\vtree^3 \!=\! \vtree^1$ and $\beta = \set{((\vtree^4_l, 0, \emp),\! (\vtree^4_r, \vtree^2, \alpha))}$.
\end{enumerate}

\begin{algorithm}[!t]
	\smaller
	\caption{{\tt Apply}$(F, G, \circ)$}
	\label{alg:apply}
	\SetKwInOut{Input}{Input}
	\SetKwInOut{Output}{Output}
	\Input{$F$: a STSDD $(\vtree^1, \vtree^2, \alpha)$; \\
		$G$: a STSDD $(\vtree^3, \vtree^4, \beta)$; \\ 
		$\circ$: an operator on combination sets ($\cap$, $\cup$ or $\setminus$).}
	\Output{$H$: The resulting STSDD $(\vtree^5, \vtree^6, \gamma)$.}
	
	\lIf{Some cases are satisfied}{
		\Return{predefined results}		
	}\vspace*{-1mm}
	
	\lIf{${\tt Cache}(F, G, \circ) \neq nil$}{
		\Return{${\tt Cache}(F, G, \circ)$}
	}\vspace*{-1mm}
	
	\uIf{$\vtree^1$ and $\vtree^3$ are incomparable}
	{
		$\vtree^5 \assign {\tt Lca}(\vtree^1, \vtree^3)$ and $\vtree^6 \assign \vtree^5$ \\		
		$F' \assign {\tt Normalize1}(F, \vtree^5)$ and $G' \assign {\tt Normalize1}(G, \vtree^5)$
	}
	\uElseIf{$\vtree^1 \propSubtree \vtree^3$}
	{
		$\vtree^5 \assign \vtree^3$	and	$\vtree^6 \assign \vtree^3$ \\		
		$F' \assign {\tt Normalize1}(F, \vtree^5)$ and $G' \assign {\tt Normalize2}(G, \vtree^6)$
	}
	\uElseIf{$\vtree^3 \propSubtree \vtree^1$}
	{
		$\vtree^5 \assign \vtree^1$	and	$\vtree^6 \assign \vtree^1$ \\
		$F' \assign {\tt Normalize2}(F, \vtree^6)$ and $G' \assign {\tt Normalize1}(G, \vtree^5)$
	}
	\Else
	{
		$\vtree^5 \assign \vtree^1$	and	$\vtree^6 \assign {\tt Lca}(\vtree^2, \vtree^4)$ \\		
		$F' \assign {\tt Normalize2}(F, \vtree^6)$ and $G' \assign {\tt Normalize2}(G, \vtree^6)$
	}\vspace*{-1mm}
	
	$\gamma \assign \emptyset$
	
	\ForEach{element $(p_i, s_i)$ of $F'$} {
		\ForEach{element $(q_j, r_j)$ of $G'$} {
			$p \assign {\tt Apply}(p_i, q_j, \cap)$\\
			\If{$\stdExtSem{p} \neq \emptyset$} {
				$s \assign {\tt Apply}(s_i, r_j, \circ)$\\
				add element $(p, s)$ to $\gamma$\\
			}\vspace*{-1mm}
		}\vspace*{-1mm}
	}\vspace*{-1mm}
	
	$H \assign {\tt Trim}({\tt Compress}(\vtree^5, \vtree^6, \gamma)))$
	
	${\tt Cache}(F, G, \circ) \assign H$
	
	\Return{$H$}
	\vspace*{-1mm}
\end{algorithm}

The {\tt Apply} algorithm, illustrated in Algorithm \ref{alg:apply}, aims to compute the binary operation $\circ$ on two STSDDs $F: (\vtree^1, \vtree^2, \alpha)$ and $G: (\vtree^3, \vtree^4, \beta)$ where $\circ$ is one of the three operations on combination sets: intersection ($\cap$), union ($\cup$) or difference ($\setminus$).
For some simple cases, we can directly return the predefined results (line 1).
For example, if $F = (0, 0, \false)$ and $\circ = \cup$, then the resulting STSDD $H$ is $G$.
Now we consider the case where $\alpha$ and $\beta$ are decomposition nodes.
In general, $\vtree^1 \neq \vtree^2$ and $\vtree^3 \neq \vtree^4$.
Therefore, it is necessary to convert $F$ and $G$ into their equivalent STSDD $F'$ and $G'$ with the same primary vtree $\vtree^5$ and secondary vtree $\vtree^6$ via normalization rules (lines 3 -- 14).
If $\vtree^1$ and $\vtree^3$ are incomparable, then both $\vtree^5$ and $\vtree^6$ are the least common ancestor of $\vtree^1$ and $\vtree^3$.
The transformed STSDDs $F'$ and $G'$ can be obtained via the first type of normalization rules.
The other cases can be handled similarly.
Let $F' = (\vtree^5, \vtree^6, \set{(p_1, s_1), \cdots, (p_n, s_n)})$ and $G' = (\vtree^5, \vtree^6, \set{(q_1, r_1), \cdots, (q_m, r_m)})$.
It is easily verified that  $H = (\vtree^5, \vtree^6, \gamma)$ where $\gamma = \set{(p_i \cap q_j, s_i \circ r_j) \mid 1 \leq i \leq n \text{ and } 1 \leq j \leq m \text{ and } p_i \cap q_j \neq \emptyset}$ (lines 16 -- 21).
Finally, compressing and trimming rules will be performed on $H$ to gain the canonicity property (line 22).
In addition, we use the cache table to avoid the recomputation on the same TSDDs and operation (lines 2 \& 23).
Let $n$ be the number of subvtrees of $\vtree^5$, and $|\alpha|$ and $|\beta|$ the size of $\alpha$ and $\beta$, respectively.
The {\tt Apply} algorithm runs in $O(n \cdot |\alpha| \cdot |\beta|)$ without the compression rules.
When we consider compressing TSDDs, the time complexity is exponential in $|\alpha|$ and $|\beta|$ in the worst case.
The above time complexity result of the {\tt Apply} algorithm still holds for SDDs.
It however was demonstrated in \cite{BroD2015} that compiling any combination set into compressed SDDs is significantly more efficient than without compressed SDDs.
The application of compression rules results in a canonical form of SDDs and hence stipulating that no two SDDs representing the same combination set are stored in the unique table, and facilitating caching in practice.
As an extension to SDDs, TSDDs have many characteristics in common with SDDs.
We focus on only compressed TSDDs in the remaining of this paper.

\begin{algorithm}[!t]
	\smaller
	\caption{{\tt OrthogonalJoin}($F$, $G$)}
	\label{alg:orthJoin}
	\SetKwInOut{Input}{Input}
	\SetKwInOut{Output}{Output}
	\Input{$F$: a STSDD $(\vtree^1,\vtree^2, \alpha)$; \\
		$G$: a STSDD $(\vtree^3, \vtree^4, \beta)$; \\ 
	}
	\Output{$H$: The resulting STSDD.}
	
	\lIf(\tcc*[f]{$\stdExtSem{F} = \emptyset$ or $\stdExtSem{G} = \emptyset$}){$F = (0, 0, \false)$ or $G = (0, 0, \false)$} 
	{
		\Return{$(0, 0, \false)$}
	}
	
	\lIf(\tcc*[f]{$\stdExtSem{F} = \set{\emptyset}$}){$F = (0, 0, \emp)$} 
	{
		\Return{$G$}
	}
	
	\lIf(\tcc*[f]{$\stdExtSem{G} = \set{\emptyset}$}){$G = (0, 0, \emp)$}
	{
		\Return{$F$} 
	}
	
	$\vtree \leftarrow$ the least common ancestor of $\vtree^1$ and $\vtree^3$
	
	\If{$\vtree^1 \subtree \vtree_l$}
	{
		$\comp{F} \leftarrow$ an STSDD s.t. $\stdExtSem{\comp{F}} = \univSet_{\vtree_l} \setminus \stdExtSem{F}$ \\		
		$H \leftarrow {\tt Trim}((\vtree, \vtree, \set{(F, G), (\comp{F}, (0, 0, \false))}))$
	}\vspace*{-1mm}
	\Else
	{
		$\comp{G} \leftarrow$ an STSDD s.t. $\stdExtSem{\comp{G}} = \univSet_{\vtree_l} \setminus \stdExtSem{G}$ \\
		$H \leftarrow {\tt Trim}((\vtree, \vtree, \set{(G, F), (\comp{G}, (0, 0, \false))}))$
	}\vspace*{-1mm}
	
	\Return{$H$}
	\vspace*{-1mm}
\end{algorithm}

\begin{algorithm}[!t]
	\smaller
	\caption{{\tt Change}($F$, $x$)}
	\label{alg:change}
	\SetKwInOut{Input}{Input}
	\SetKwInOut{Output}{Output}
	\Input{$F$: a STSDD $(\vtree^1,\vtree^2, \alpha)$; \\
		$x$: a variable.}
	\Output{$G$: The resulting STSDD.}
	
	$\vtree^3 \leftarrow$ the leaf node labeled by $x$ \\
	
	\lIf(\tcc*[f]{$\stdExtSem{F} = \set{\emptyset}$}){$F = (0, 0, \emp)$}
	{
		\Return{$(\vtree^3, \vtree^3, \negempty)$}
	}
	
	\lIf(\tcc*[f]{$\stdExtSem{F} = \set{\set{x}}$}){$F = (\vtree^3, \vtree^3, \negempty)$}
	{
		\Return{$(0, 0, \emp)$}
	}
	
	\lIf(\tcc*[f]{$\stdExtSem{F} = \emptyset$ or $\stdExtSem{F} = \set{\set{x}, \emptyset}$}){$F = (0, 0, \false)$ or $F = (\vtree^3, 0, \emp)$}
	{
		\Return{$F$}
	}
	
	\lIf{$\vtree^3 \propSubtree \vtree^1$ {\bf and} $\vtree^3$ is not a subvtree of $\vtree^2$}{
		\Return{$F$}
	}
	\lIf{${\tt Cache}(F, x, {\tt Change}) \neq nil$}{
		\Return{${\tt Cache}(F, x, {\tt Change})$}
	}
	
	\If{$\vtree^1$ and $\vtree^3$ are incomparable}{
		$G \leftarrow {\tt OrthogonalJoin}(F, (\vtree^3, \vtree^3, \negempty))$
	}\vspace*{-1mm}
	\ElseIf{$\vtree^3 = \vtree^2$}{
		$\vtree^2_p \leftarrow$ the parent node of $\vtree^2$
		
		\If{$\vtree^2 = (\vtree^2_p)_l$}{
			$H \leftarrow$ an STSDD s.t. $\stdExtSem{H} = \univSet_{(\vtree^2_p)_l} \setminus \set{\emptyset}$
			
			$G \leftarrow (\vtree^1, \vtree^2_p, \set{((0, 0, \emp),((\vtree^2_p)_r, 0, \emp)), (H, (0, 0, \false))})$
		} \vspace*{-1mm}
		\Else(\tcc*[f]{$\vtree^2 = (\vtree^2_p)_r$}){
			$G \leftarrow (\vtree^1, \vtree^2_p, \set{(((\vtree^2_p)_l, 0, \emp), (0, 0, \emp))})$
		}\vspace*{-1mm}
	}\vspace*{-1mm}
	\Else(\tcc*[f]{$\vtree^3 \propSubtree \vtree^2$}){
		$\gamma \leftarrow \emptyset$\\
		\ForEach{element $(p_i, s_i)$ of $\alpha$} {
			\If{$\vtree^3 \subtree \vtree^2_l$}{
				add element $({\tt Change}(p_i,X), s_i)$ to $\gamma$
			}\vspace*{-1mm}
			\Else(\tcc*[f]{$\vtree^3 \subtree \vtree^2_r$}){
				add element $(p_i, {\tt Change}(s_i,X))$ to $\gamma$
			}\vspace*{-1mm}
		}\vspace*{-1mm}
		$G \leftarrow (\vtree^1, \vtree^2, \gamma)$
	}\vspace*{-1mm}
	
	$G \leftarrow {\tt Trim}(G)$
	
	${\tt Cache}(F, x, {\tt Change}) \leftarrow G$
	
	\Return{$G$}
	\vspace*{-1mm}
\end{algorithm}

\begin{table*}
	\vspace*{-4mm}
	\tiny
	\centering
	\caption{The comparison among SDDs, ZSDDs, NSTSDDs, NZTSDDs, ESTSDDs and EZTSDDs over $4$ categories of benchmarks}
	\label{tab:result}	
	%\vspace*{-4mm}
	\begin{tabular}{|c|rr|rr|rr|rr|rr|rr|r|}
		\hline
		Benchmarks    & \multicolumn{2}{c|}{SDD}    & \multicolumn{2}{c|}{ZSDD}    & \multicolumn{2}{c|}{NSTSDD}    & \multicolumn{2}{c|}{NZTSDD}     & \multicolumn{2}{c|}{ESTSDD}   & \multicolumn{2}{c|}{EZTSDD} & \\

		& size  & time               & size  & time                  & size  & time                & size  & time                     & size  & time               & size  & time & best size\\ \hline
		Words-Binary-Compact   & 1,562,787 & 100.735    & 922,405 & 34.884    & 935,605 & 23.289    & \bf{636,393} & 41.198    & 937,770 & 41.463    & 1,005,408 & 60.207   & \bf{636,393}
		\\
		Words-Binary-ASCII   & 2,356,875 & 110.015    & 1,610,952 & 64.302    & 1,602,317 & 38.075    & \bf{802,562} & 65.870    & 1,683,324 & 78.939    & 1,780,959 & 98.862  & \bf{802,562} 
		\\		
		Words-OneHot-Compact   & 19,205,734 & 755.775    & 593,762 & 26.218    & \bf{593,624} & 18.800    & 706,927 & 68.202    & 593,766 & 28.809    & 706,968 & 86.394   & \bf{593,624}
		\\
		Words-OneHot-ASCII   & 46,281,905 & 2,472.188    & 594,124 & 27.261    & \bf{594,100} & 16.834    & 707,912 & 61.939    & 594,126 & 31.508    & 707,914 & 80.362   & \bf{594,100}
		\\
		Passwords-Binary-Compact   & - & -    & 3,488,141 & 339.656    & \bf{3,440,014} & 369.770    & 3,464,514 & 437.667    & 3,488,592 & 755.580    & 3,834,894 & 767.570   & \bf{3,440,014}
		\\
		Passwords-Binary-ASCII   & - & -    & 5,664,643 & 764.867    & \bf{5,579,382} & 711.813    & 5,733,694 & 721.128    & 5,732,059 & 1,337.874    & 6,061,272 & 1,336.517   & \bf{5,579,382}
		\\
		Passwords-OneHot-Compact   & - & -    & 2,243,018 & 266.236    & \bf{2,243,016} & 265.143    & 2,663,610 & 1,620.616    & 2,243,018 & 457.473    & 2,663,610 & 1,965.936   & \bf{2,243,016}
		\\
		Passwords-OneHot-ASCII   & - & -    & \bf{2,258,146} & 257.638    & 2,258,216 & 284.455    & 2,681,843 & 1,052.822    & \bf{2,258,146} & 512.400    & 2,681,843 & 1,190.616   & \bf{2,258,146}
		\\
		\hline
		8-Queens-Binary   & 1,336 & 0.045    & 858 & 0.092    & \bf{830} & 0.083    & 960 & 0.065    & 854 & 0.068    & 952 & 0.07    & \bf{830}
		\\
		9-Queens-Binary   & 4,696 & 0.185    & 3,001 & 0.311    & \bf{2,897} & 0.332    & 3,398 & 0.193    & 3,052 & 0.302    & 3,422 & 0.216   & \bf{2,897}
		\\
		10-Queens-Binary   & \bf{6,017} & 2.263    & 6,839 & 1.013    & 6,222 & 0.664    & 6,568 & 1.617    & 7,450 & 0.794    & 8,168 & 0.692   & 6,222
		\\
		11-Queens-Binary   & 32,835 & 6.668    & 21,314 & 30.874    & 19,441 & 2.246    & 20,453 & 8.654    & 19,826 & 140.624    & \bf{19,123} & 21.130   & \bf{19,123}
		\\
		12-Queens-Binary   & 118,925 & 30.833    & 102,345 & 26.736    & \bf{81,192} & 11.592    & 117,215 & 20.406    & 97,545 & 2,247.166    & 89,546 & 54.305   & \bf{81,192}
		\\
		13-Queens-Binary   & 666,910 & 52.210    & 433,856 & 5,061.758    & \bf{35,2637} & 160.228    & 511,928 & 124.993    & 466,128 & 1,171.961    & 48,6257 & 241.812   & \bf{35,2637}
		\\
		14-Queens-Binary   & 2,610,502 & 556.376    & 1,871,878 & 5,943.739    & \bf{1,554,825} & 682.726    & 2,340,878 & 450.193    & 2,134,480 & 511.551    & 2,371,137 & 366.242  & \bf{1,554,825} 
		\\
		%		Queen-Binary-15   & 14,642,804 & 833.623    & 10,691,180 & 3,025.764    & \bf{8,176,783} & 6,397.268    & 11,984,251 & 916.124    & 10,982,657 & 1,392.373    & 12,148,699 & 1,164.52   
		%		\\
		%		average		& 2,260,503.123 & 185.275		& 1,641,408.875 & 1,761.286	& 1,274,353.375 & 906.892		& 1,873,206.375 & 190.281		& 1,713,999 & 683.105		& 1,890,913 & 231.123 \\
		8-Queens-OneHot   & 2,691 & 0.178    & 953 & 0.266    & 763 & 0.167    & 832 & 0.144    & \bf{730} & 0.147    & 837 & 0.208   & \bf{730}
		\\
		9-Queens-OneHot   & 12,526 & 0.359    & 3,213 & 0.549    & 2,616 & 0.344    & 3,170 & 0.301    & \bf{2,604} & 0.349    & 3,171 & 0.338 & \bf{2,604}  
		\\
		10-Queens-OneHot   & 24,542 & 0.903    & 7,213 & 1.517    & 6,677 & 3.525    & 7,462 & 1.368    & \bf{6,520} & 4.481    & 7,467 & 1.099   & \bf{6,520}
		\\
		11-Queens-OneHot   & 95,390 & 4.049    & 22,729 & 15.659    & \bf{21,340} & 19.492    & 25,780 & 4.614    & 21,585 & 138.004    & 26,015 & 40.781   & \bf{21,340}
		\\
		12-Queens-OneHot   & 376,669 & 77.366    & 98,148 & 168.901    & 94,134 & 798.095    & 100,139 & 182.361    & \bf{91,622} & 798.389    & 115,873 & 117.148  & \bf{91,622} 
		\\
		13-Queens-OneHot   & 2,096,517 & 231.321    & 431,436 & 92.528    & \bf{409,525} & 677.2    & 491,912 & 279.847    & 416,729 & 974.899    & 523,479 & 278.24   & \bf{409,525}
		\\
		14-Queens-OneHot   & 8,604,470 & 623.433    & 1,923,847 & 4,313.754    & 1,863,942 & 3,102.857    & 2,348,386 & 620.163    & \bf{1,822,922} & 1,111.114    & 2,348,381 & 603.111   & \bf{1,822,922}
		\\
		%average		& 1,601,829.286 & 133.944		& 355,362.714 & 656.168	& 342,713.857 & 657.383		& 425,383 & 155.543		& 337,530.286 & 432.483		& 432,174.714 & 148.704 \\
		\hline
		AutoFlight-PT-06a   & 2,276 & 167.958    & - & -    & \bf{855} & 5060.282    & 1,228 & 254.188    & 889 & 7,185.497   & - & -	& \bf{855}
		\\
		%		BusinessProcesses-PT-04   & - & -    & - & -    & - & -    & - & -    & - & -   & - & - 
		%		\\
		Dekker-PT-015   & 606 & 5.094    & 3,562 & 216.347    & 408 & 7.699    & \bf{393} & 5.665    & 418 & 7.349    & 399 & 5.949   & \bf{393}
		\\
		%		DES-PT-01b   & - & -    & - & -    & - & -    & - & -    & - & -   & - & - 
		%		\\
		DiscoveryGPU-PT-14a   & 1,188 & 3,758.586    & - & -    & \bf{372} & 627.382    & - & -    & 496 & 1,681.220    & 739 & 6,228.986   & \bf{372}
		\\
		DLCround-PT-04a   & 1,016 & 49.017    & 734 & 742.783    & \bf{463} & 157.861    & 547 & 16.599    & 514 & 335.422    & 571 & 16.350   & \bf{463}
		\\
		FlexibleBarrier-PT-12a   & - & -    & - & -    & \bf{445} & 2,447.236    & 732 & 2,924.438    & 474 & 2,282.550   & - & -	& \bf{445}
		\\
		IBM319-PT-none   & 3,013 & 3.606    & \bf{726} & 16.185    & 837 & 19.177    & 810 & 11.799    & 873 & 9.773   & - & - & 810
		\\
		LamportFastMutEx-PT-4   & 43,597 & 418.267    & - & -    & 11,087 & 1,131.252    & 24,759 & 614.258    & \bf{4,241} & 843.778    & 7,210 & 204.395   & \bf{4,241}
		\\
		MAPKbis-PT-5320   & 558 & 64.483    & 771 & 1,997.135    & - & -    & 415 & 86.565    & - & -    & \bf{372} & 66.46   & \bf{372}
		\\
		NeoElection-PT-3   & 6,039 & 29.293    & - & -    & 2,395 & 3,800.556    & 1,908 & 148.984    & - & -    & \bf{1,882} & 83.911   & \bf{1,882}
		\\
		NQueens-PT-08   & 243,080 & 170.705    & - & -    & 229,506 & 2,136.447    & 271,281 & 173.608    & \bf{193,757} & 1,290.575    & 316,199 & 240.006   & \bf{193,757}
		\\
		ParamProductionCell-PT-4   & 322,645 & 1,214.601    & - & -    & - & -    & \bf{7,755} & 1,764.348    & - & -    & 20,939 & 797.247   & \bf{7,755}
		\\
		Peterson-PT-2   & 3,477 & 4.727    & 1,821 & 24.098    & 2,207 & 38.392    & \bf{1,128} & 9.160    & 1,737 & 48.345    & 1,164 & 19.055   & \bf{1,128}
		\\
		Philosophers-PT-000010   & 1,330 & 2.420    & 755 & 114.018    & 6,165 & 39.386    & 496 & 1.365    & \bf{221} & 7.882    & 505 & 1.540   & \bf{221}
		\\
		Raft-PT-03   & 375 & 1.955    & 432 & 2.423    & \bf{234} & 1.314    & 292 & 1.937    & 239 & 1.462    & 314 & 1.508    & \bf{234}
		\\
		Railroad-PT-010   & 2,462 & 6.619    & 16,033 & 694.707    & \bf{1,191} & 20.191    & 1,249 & 8.802    & 1,352 & 8.949   & - & - & \bf{1,191}
		\\
		ResAllocation-PT-R020C002   & 1,876 & 228.02    & - & -    & 1,334 & 2,935.194    & 1,162 & 658.513    & 965 & 1,564.616    & \bf{808} & 186.662   & \bf{808}
		\\
		Referendum-PT-020   & 413 & 1.598    & 267 & 328.033    & \bf{173} & 1.497    & 258 & 3.005    & 238 & 1.689    & 224 & 1.458   & \bf{173}
		\\
		RwMutex-PT-r0020w0010   & 829 & 0.918    & 797 & 2.007    & \bf{455} & 4.042    & 586 & 1.377    & 52,032 & 5,796.452    & 465 & 1.831   & \bf{455}
		\\
		SafeBus-PT-03   & 3,113 & 1.525    & 6,871 & 7.664    & 1,565 & 1.829    & 2,152 & 2.520    & \bf{1,360} & 2.596    & 1,543 & 2.104   & \bf{1,360}
		\\
		SharedMemory-PT-000010   & 2,339 & 3.768    & 1,746 & 11.448    & 809 & 3.592    & 1,414 & 6.219    & \bf{680} & 3.438    & 739 & 10.231  & \bf{680} 
		\\
		SimpleLoadBal-PT-10   & 16,136 & 1,040.229    & - & -    & - & -    & \bf{5,365} & 499.745    & - & -    & 85,566 & 5,353.878   & \bf{5,365}
		\\
		\hline		
		9symml   & 27,119 & 80.097    & 22,152 & 2,102.474    & 19,729 & 23.757    & 24,399 & 120.719    & 22,520 & 585.506    & \bf{16,442} & 100.958   & \bf{16,442}
		\\
		alu2   & - & -    & - & -    & - & -    & 13,658 & 573.479    & - & -    & \bf{13,208} & 1,219.279   & \bf{13,208}
		\\
		apex6   & - & -    & - & -    & - & -    & \bf{1,166,371} & 5,573.009    & - & -   & - & - & \bf{1,166,371}
		\\
		apex7   & 7,800 & 13.541    & 8,688 & 17.721    & 20,022 & 173.013    & \bf{5,958} & 18.721    & - & -    & 12,972 & 37.326    & \bf{5,958}
		\\
		b9   & \bf{8,096} & 9.849    & 19,066 & 121.499    & 24,115 & 696.112    & 18,192 & 18.052    & 69,642 & 210.751    & 12,719 & 20.033   & 12,719
		\\
		C432   & \bf{11,190} & 9.503    & - & -    & - & -    & 12,990 & 205.840    & 17,113 & 101.410    & 30,777 & 69.729   & 12,990
		\\
		C499   & \bf{294,686} & 1,560.059    & - & -    & - & -    & - & -    & - & -   & - & -		& -
		\\
		c8   & 19,656 & 49.683    & 66,069 & 972.876    & 19,681 & 177.154    & 15,849 & 67.920    & \bf{14,644} & 29.474    & 17,382 & 76.453  & \bf{14,644}
		\\
		%		cc   & 2,054 & 1.304    & 3,245 & 2.413    & 2,412 & 1.996    & \bf{1,434} & 2.046    & 1,679 & 50.534    & 2,408 & 2.549   & \bf{1,434}		\\
		%		cht   & 4,146 & 3.510    & 4,843 & 6.571    & 4,035 & 8.223    & 4,848 & 4.337    & \bf{3,615} & 3.138    & 4,638 & 8.401    & \bf{3,615}		\\
		%		cm150a   & 1,622 & 0.615    & 2,110 & 1.725    & 2,790 & 0.780    & 1,695 & 1.201    & \bf{1,491} & 0.651    & 2,295 & 1.456   & \bf{1,491}		\\
		%cm85a   & 1,328 & 0.424    & 2,483 & 1.337    & 1,062 & 1.350    & \bf{869} & 1.062    & 998 & 0.436    & 959 & 0.878   & \bf{869}		\\
		%		cmb   & \bf{1,288} & 0.673    & 1,548 & 1.141    & 1,528 & 1.173    & 2,337 & 0.677    & 1,699 & 0.717    & 1,548 & 1.197   & 1,528		\\
		%		comp   & 2,549 & 2.289    & \bf{2,497} & 4.887    & 2,694 & 5.353    & 3,170 & 5.284    & 2,746 & 2.195    & 3,077 & 3.811   & 2,694		\\
		count   & 4,027 & 2.018    & 4,126 & 3.008    & 5,175 & 6.823    & 6,051 & 7.400    & 199,478 & 248.969    & \bf{2,864} & 6.569   & \bf{2,864}
		\\
		%		cu   & 2,038 & 0.819    & 1,548 & 0.992    & 2,495 & 2.777    & \bf{1,437} & 1.141    & 2,309 & 0.782    & 1,916 & 1.736   & \bf{1,437} 		\\
		example2   & 9,246 & 18.161    & 9,925 & 54.859    & \bf{8,597} & 204.963    & 11,172 & 31.125    & 19,650 & 27.535    & 18,079 & 64.686   & \bf{8,597}
		\\
		%		f51m   & 4,940 & 1.089    & 5,374 & 7.725    & \bf{2,725} & 10.252    & 2,787 & 2.394    & 5,078 & 4.490    & 3,216 & 5.593   & \bf{2,725}		\\
		frg1   & 137,905 & 60.219    & - & -    & - & -    & 200,455 & 1,501.116    & 190,457 & 3,129.121    & \bf{121,959} & 638.711   & \bf{121,959}
		\\
		lal   & 8,540 & 4.178    & 8,172 & 48.631    & 40,712 & 266.991    & \bf{5,978} & 9.787    & 12,061 & 8.583    & 75,768 & 40.378   & \bf{5,978}
		\\
		%		ldd   & 2,085 & 1.497    & 1,834 & 3.442    & 1,498 & 3.990    & 920 & 2.270    & 1036 & 2.532    & \bf{844} & 5.897   & \bf{844}		\\
		mux   & 2,071 & 0.884    & 8,588 & 6.445    & 5,272 & 3.578    & \bf{1,718} & 1.701    & 1,788 & 0.895    & 3,339 & 1.995   & \bf{1,718}		\\
		%		my\_adder   & 3,278 & 2.652    & 3,966 & 3.603    & 3,703 & 4.018    & 2,753 & 3.508    & 2,217 & 3.474    & \bf{2,148} & 6.327   & \bf{2,148} 	\\
		%		pm1   & 2,431 & 1.190    & 2,701 & 3.967    & 2,659 & 1.389    & \bf{2,165} & 1.500    & 3,447 & 1.293    & 2,470 & 2.559   & \bf{2,165} 		\\
		sct   & 8,078 & 6.147    & 71,845 & 90.752    & 39,136 & 89.677    & \bf{7,529} & 9.910    & 15,638 & 11.557    & 7,777 & 9.230   & \bf{7,529}
		\\
		term1   & \bf{100,589} & 371.290    & - & -    & - & -    & 126,324 & 554.553    & - & -    & 3,500,493 & 4,524.447   & 126,324
		\\
		ttt2   & 24,619 & 53.182    & - & -    & 29,290 & 526.672    & \bf{17,382} & 123.076    & 19,868 & 201.311    & 32,908 & 224.050   & \bf{17,382}
		\\
		unreg   & 3,973 & 2.256    & 4,178 & 2.297    & 8,188 & 7.457    & \bf{3,782} & 3.724    & 24,445 & 11.645    & 3,850 & 4.620   & \bf{3,782}
		\\
		vda   & - & -    & - & -    & - & -    & \bf{17,338} & 652.971    & - & -   & - & -	& \bf{17,338}
		\\
		x4   & 36,364 & 272.750    & - & -    & - & -    & \bf{24,667} & 316.670    & - & -    & 25,358 & 439.805   & \bf{24,667}
		\\
		%		z4ml   & 2,725 & 0.568    & 1,422 & 1.450    & 2,218 & 1.628    & 1,915 & 0.833    & 1,116 & 0.591    & \bf{980} & 1.248   & \bf{980}		\\
		c1908   & \bf{85,725,495} & 5,324.091    & - & -    & - & -    & - & -    & - & -   & - & -		& -
		\\
		c432   & \bf{11,647} & 25.205    & - & -    & - & -    & 19,277 & 21.255    & 126,779 & 1516.911    & 117,939 & 117.537   & 19,277
		\\
		c499   & \bf{131,517} & 29.554    & - & -    & - & -    & 245,306 & 81.118    & 307,962 & 297.176    & 160,305 & 70.929   & 160,305
		\\
		s1196   & - & -    & - & -    & - & -    & - & -    & - & -    & \bf{174,796} & 1631.269   & \bf{174,796}
		\\
		s1238   & \bf{96,122} & 1,169.499    & - & -    & - & -    & - & -    & - & -   & - & -		& -
		\\
		s1494   & - & -    & - & -    & - & -    & - & -    & - & -    & \bf{210,342} & 3,330.731   & \bf{210,342}
		\\
		%		s208.1   & 2,023 & 1.310    & 2,220 & 2.129    & 2,386 & 2.939    & 2,096 & 1.731    & 2,270 & 1.251    & \bf{1,694} & 2.303    & \bf{1,694}		\\
		s298   & 3,864 & 2.860    & 4,177 & 2.326    & 9,880 & 87.201    & \bf{3,631} & 5.504    & 4,979 & 5.565    & 4,110 & 3.872   & \bf{3,631}		\\
		s344   & \bf{3,819} & 2.676    & 5,047 & 5.611    & 7,114 & 11.225    & 4,380 & 4.612    & 6,285 & 8.446    & 4,296 & 5.904   & 4,296		\\
		s349   & 4,291 & 3.715    & 5,760 & 3.233    & 7,047 & 10.201    & 9,364 & 6.520    & \bf{3,856} & 7.049    & 6,783 & 7.731   & \bf{3,856} 	\\
		s382   & 5,326 & 3.631    & 6,017 & 7.018    & 5,725 & 8.781    & \bf{3,374} & 11.541    & - & -    & 4,182 & 7.822   & \bf{3,374}
		\\
		s386   & \bf{8,920} & 5.996    & 25,225 & 85.237    & 78,638 & 892.482    & 10,193 & 9.641    & 27,644 & 49.346    & 9,369 & 10.551   & 9,369
		\\
		s400   & \bf{4,319} & 4.152    & 5,988 & 5.581    & 5,397 & 10.359    & 4,694 & 22.990    & 13,059 & 9.187    & 40,038,280 & 3,738.312   & 4,694
		\\
		s420   & \bf{4,888} & 5.019    & 9,711 & 13.000    & 5,958 & 11.267    & 18,836 & 24.126    & 10,050 & 165.934    & 5,030 & 7.952   & 5,030
		\\
		s444   & 4,955 & 3.894    & 6,656 & 6.771    & 4,359 & 5.135    & 4,615 & 16.479    & 7,049 & 235.753    & \bf{3,898} & 6.436   & \bf{3,898}		\\
		s510   & 21,173 & 16.945    & - & -    & - & -    & \bf{13,440} & 15.398    & - & -    & 23,496 & 57.638   & \bf{13,440}
		\\
		s526N   & 8,247 & 10.277    & - & -    & - & -    & \bf{7,088} & 16.894    & - & -   & - & -	& \bf{7,088}
		\\
		s526   & 7,100 & 6.656    & 31,253 & 459.766    & 14,385 & 69.232    & \bf{6,361} & 7.607    & - & -    & 6,891 & 18.935   & \bf{6,361}
		\\
		s641   & \bf{11,581} & 12.924    & 51,062 & 113.329    & 21,219 & 103.634    & 18,208 & 384.775    & 71,137 & 72.105    & 16,758 & 24.438   & 16,758
		\\
		s713   & \bf{20,526} & 11.598    & 56,896 & 773.796    & 45,448 & 371.679    & 33,829 & 29.134    & - & -    & 27,558 & 28.119   & 27,558
		\\
		s832   & 28,080 & 59.900    & - & -    & - & -    & 31,081 & 704.611    & - & -    & \bf{21,719} & 175.599   & \bf{21,719}
		\\
		s838.1   & 10,108 & 24.724    & 12,307 & 56.800    & 17,204 & 71.095    & \bf{8,136} & 49.535    & - & -   & - & -	& \bf{8,136}
		\\
		s838   & \bf{13,493} & 30.427    & 22,807 & 54.466    & 15,442 & 124.552    & 15,836 & 204.722    & - & -    & 23,702 & 108.166   & 15,442
		\\
		s953   & \bf{119,689} & 326.609    & - & -    & - & -    & - & -    & - & -    & 251,112 & 1,564.516   & 251,112 	\\
		\hline
		%Avg.        & 770           & 7     & 9841          & 12757          & 99    & 784750          & 12757          & 99    & 784750         & 12650           & 150    & 655686            & 9128          & 574   & 283668         \\ \hline
	\end{tabular}
	\vspace*{-2mm}
\end{table*}

\subsection{Orthogonal Join and Change}
%The join operation is 
We begin by introducing the algorithm for orthogonal join, illustrated in Algorithm \ref{alg:orthJoin}.
Assume that we are given two STSDDs $F = (\vtree^1, \vtree^2, \alpha)$ and $G = (\vtree^3, \vtree^4, \beta)$.
Algorithm \ref{alg:orthJoin} requires that $F$ and $G$ are orthogonal, that is, $\vtree^1$ and $\vtree^3$ are incomparable.
%It works on the premise that $F$ and $G$ are orthogonal.
The resulting STSDD is also the empty set, if one of the STSDDs $F$ and $G$ is the empty set $\emptyset$ (line 1).
In the case where $F$ (resp. $G$) denotes the combination set $\set{\emptyset}$, the outcome is $G$ (resp. $F$) (lines 2 \& 3).
In general, we let $\vtree$ be the least common ancestor of $\vtree^1$ and $\vtree^3$.
If $\vtree^1$ is a subvtree of the left child of $\vtree$, then the result STSDD $H$ is $(\vtree, \vtree, \set{(F,G), (\comp{F}, (0, 0, \false))})$ where $\stdExtSem{\comp{F}} = \univSet_{\vtree_l} \setminus \stdExtSem{F}$ (lines 5 -- 7).
The opposite direction can be similarly handled (lines 8 -- 10).
Algorithm \ref{alg:orthJoin} runs in a constant time.

\looseness=-1
Another basic operation for combination set is change.
It takes a STSDD $F$ and a variable $x$ as inputs, and outputs a STSDD $G$ s.t. $\stdExtSem{G} = \change(\stdExtSem{F}, x)$.
Let $\vtree^3$ be the leaf vtree node with the label $x$.
We first consider three special cases.
If $F$ denotes the combination set $\set{\emptyset}$, then the resulting STSDD $G$ represents $\set{\set{x}}$, and vice versa (lines 2 \& 3).
If one of the following three cases hold: (1) $F = (0, 0, \false)$; or (2) $F=(\vtree^3, 0, \emp)$; or (3) $\vtree^3 \prec \vtree^1$ and $\vtree^3$ is not a subtree of $\vtree^2$, then the change operation do not modify the input STSDD $F$ (lines 4 \& 5).
In the case where $\vtree^1$ and $\vtree^3$ are incomparable, then the change of $F$ by $x$ is the orthogonal join of the two STSDDs $F$ and $(\vtree^3, \vtree^3, \negempty)$.
If none of the above cases holds, then $\vtree^3 \subtree \vtree^2$.
We analyze the following two cases: $\vtree^3 = \vtree^2$ and $\vtree^3 \propSubtree \vtree^2$.
In the case where $\vtree^3 =\vtree^2$, we construct $G$ as $(\vtree^1, \vtree^2_p, \set{((0, 0, \emp), ((\vtree^2_p)_r, 0, \emp)), (H, (0, 0, \false))})$ where $H$ denotes the complement of $\set{\emptyset}$ if $\vtree^2$ is the left child of its parent $\vtree^2_p$ (lines 9 -- 13); and as $(\vtree^1, \vtree^2_p, \set{(((\vtree^2_p)_l, 0, \emp), (0, 0, \emp))})$ if $\vtree^2$ is the right child of $\vtree^2_p$ (lines 14 \& 15).
In the case where $\vtree^3 \propSubtree \vtree^2$, $\alpha$ must be a decomposition node.
We recursively apply the change operation on the prime $p_i$ of elements of $\alpha$ if $\vtree^3 \subtree \vtree^2_l$ (lines 19 \& 20), and on the sub $s_i$ if $\vtree^3 \subtree \vtree^2_r$ (lines 21 \& 22).
Finally, we use the trimming rules on $G$ (line 24).
The algorithm uses the cache table to avoid the recomputation on the same STSDD and variable (lines 6 \& 25) and runs in linear time w.r.t. $\size{F}$.

\section{Experimental Results}

\looseness=-1
In this section, we compare four variants of TSDDs against SDDs and ZSDDs with respective to their compactness in four categories of benchmarks: dictionary, $n$-queens problems, safe petri nets and digital circuits.
For convenience, NSTSDD, NZTSDD, ESTSDD, and EZTSDD are the abbreviations for node-based STSDD, node-based ZTSDD, edge-based STSDD, and edge-based ZTSDD, respectively.
We implemented an efficient TSDD package with the proposed algorithms in C language. %based on the code of the SDD package \cite{ChoD2018}.
We have also devised minimization algorithm for TSDDs via so as to searching a good vtree and integrate this algorithm into our package because the size of TSDDs is sensitive to the vtree.
Due to the space limit, we do not present minimization algorithm in this paper, which will be clarified in future work.
All experiments were carried out on a machine equipped with an Intel Core i7-8086K 4GHz CPU and 64GB RAM.
Table \ref{tab:result} shows the experimental results of 6 decision diagrams for 85 test cases across 4 benchmark categories on size and time.
%\textcolor[rgb]{1,0,0}{
The columns ``size" denotes the size of compiled decision diagrams and ``time" the overall compilation runtime in seconds.
%}
The smallest sizes among six decision diagrams are highlighted in bold font.
The last column shows the smallest size among 4 variants of TSDDs.
The entry ``-" denotes a failed compilation due to the timeout of 2 hours.

\looseness=-1
The dictionaries we use are the English words in file {\tt /usr/shar/dict/words} on MacOS system with $235,886$ words of length up to $24$ from $54$ symbols and the password list with $979,247$ words of length up to $32$ from $79$ symbols \cite{Bry2020b}.
A dictionary will be encoded in two ways: \textit{binary} and \textit{one-hot}.
We consider two sets of symbols: \textit{the compact form} consisting of the symbols only found in the dictionary, and \textit{the ASCII form} consisting of all 128 characters.
%First of all, we consider the experimental data for the test cases over \textit{dictionary encoding}.
We can compile all of the $8$ test cases of dictionaries in ZSDDs and $4$ variants of TSDDs.
However, SDD compilation fails in the password dictionary.
In addition, ZSDDs and TSDDs have a significant size advantage over SDDs.
%\textcolor[rgb]{1,0,0}{
Especially for words in one-hot encoding, NSTSDDs and ESTSDDs have over 96\% fewer size than SDDs.
%}
This is because the zero-suppressed trimming rule has a tremendous benefit to reduce the size of decision diagrams for dictionaries.
Finally, NSTSDD performs the best in $6$ out of $8$ test cases and hence being an efficient representation for dictionaries in terms of time.

\looseness=-1
The $n$-Queens problem aims to place $n$ queens in such a manner on an $n \times n$ chessboard that no two queens can attack each other by being in the same row, column or diagonal.
%The number of feasible solutions grows exponentially as $n$ increases, hence, the size of decision diagrams also increases.
%Each solution to the $n$-queens problem can be considered as a truth assignment and all solutions can be represented by a Boolean function.
We also have two ways (one-hot and binary) to encode this problem.
Firstly, one of the variants of TSDDs is the most compact representation in 13 out of 14 test cases.
This indicates that combining two trimming rules can result in a smaller decision diagram than using one trimming rule.
Secondly, compilations in SDDs, NZTSDDs and EZTSDDs are more effective than the other three.
However, the size of SDD representation is obviously larger than others.
%\textcolor[rgb]{1,0,0}{
Specifically, $13$- and $14$-Queens problems in one-hot encoding are of size 2,096,517 and 8,604,470 that are approximately 5.12 and 4.62 times larger than NSTSDDs, respectively.
%}
Finally, ZSDD compilation is time-consuming, especially, $13$- and $14$-Queens problem in binary encoding take 5,061 and 5,943 seconds longer than NZTSDDs by high factors of 40.50 and 13.20, respectively.

\looseness=-1
Petri nets are a popular graphical modeling tool for representing and analyzing concurrent systems. 
A Petri net is safe iff there is at most one token in each one of its places.
We utilize decision diagrams to denote the set of reachable states of safe Petri net.
The Petri net benchmark comes from the 2018 Model Checking Contest ({\tt https://mcc.lip6.fr/2018/}).
As for the size, one of the variants TSDDs performs the best in $20$ out of $21$ test cases.
In particular, for the two large test cases: NQueens-PT-08 and ParamProductionCell-PT-4, the minimal size among TSDDs are 20.3\% and 97.6\% smaller than the size of SDDs.
Moreover, ZSDDs fails in compilation of the above two test cases.

\looseness=-1
The final benchmark we consider is digital circuits.
We use the test cases from the three sets of LGSynth89, iscas85, and iscas89 benchmarks that are widely used in CAD community.
We only present the test cases if (1) at least one decision diagrams is successfully created within the timeout of 2 hours, and (2) at least one decision diagrams has size of more than 5,000.
It can be observed that NZTSDD and EZTSDD together are able to compile the 5 test cases: alu2, apex6, vda, s1196 and s1494 while other decision diagrams cannot.
SDDs can successfully compile the 2 test cases: C499 and c1908 which the others fails.
Regarding on size, TSDDs are the most compact representation in 26 out of 42 test cases, with 1 cases in NSTSDD, 15 cases in NZTSDD, 2 cases in EZTSDD and 8 cases in ESTSDD.
SDDs provide the most succinct representation in 16 test cases.
ZSDD-representation is not the smallest one in any test case.
Apart from the size, we can see that TSDD compilation is comparatively slower than SDD compilation.
This is because the underlying data structure of TSDD is more complicated than SDD and the minimization algorithm for TSDD needs more time. 
However, finding a decision diagram with smaller size is of utmost importance.
First, keeping decision diagrams compact improves the effectiveness of subsequent Boolean operations.
Secondly, the size of decision diagrams is crucial for specific application.
For example, BDD-based logic synthesis generates electronic circuit of fewer size with smaller decision diagrams \cite{WilD2009,AmaGM2013}.
Therefore it is worthy to spend extra time in producing more compact decision diagrams.

\looseness=-1
In summary, we can observe that no single decision diagram dominates all categories of benchmarks in terms of size and compilation time.
TSDDs are a significant representation of Boolean functions as an important addition to SDDs and ZSDDs,

\section{Conclusion}
\looseness=-1
In this paper, we design four variants of TSDDs that mixes standard and zero-suppressed trimming rules.
We divide TSDDs into STSDDs and ZTSDDs according to the order of trimming rules.
The former requires the standard trimming rules to be applied first, and the latter requires the zero-suppressed trimming rules to be utilized first.
In addition, we provide two approaches to implementing TSDDs: node-based and edge-based.
Node-based implementation stores both primary vtree and secondary vtree in a TSDD node, and edge-based implementation store primary vtree as an edge pointing to a TSDD node.
Therefore, we obtain four different kinds of TSDDs: node-based STSDDs, node-based ZTSDDs, edge-based STSDDs and edge-based ZTSDDs.
We design their syntax and semantics and provide design three algorithms: Apply, OrthogonalJoin and Change for STSDDs to implement the corresponding operations over combination sets.
We finally conduct experiments on four benchmarks, which confirms that TSDDs are a more compact form compared to SDDs and ZSDDs.

\bibliographystyle{IEEEtran}
\bibliography{ICCAD-2023-1.bib}

\end{document}